# MAGIC: Multi-task Gaussian process for joint imputation and classification in healthcare time series

Dohyun Ku[a], Catherine D. Chong[b], Visar Berisha[c], Todd J. Schwedt[b], and Jing Li[a]*

*[a]H. Milton Stewart School of Industrial and Systems Engineering, Georgia Institute of Technology, GA, USA; [b]Department of Neurology, Mayo Clinic Arizona, AZ, USA; [c]College of Health Solutions and School of Electrical Computer and Energy Engineering, Arizona State University, AZ, USA*

**Abstract**

Time series analysis has emerged as an important tool for improving patient diagnosis and management in healthcare applications. However, these applications commonly face two critical challenges: time misalignment and data sparsity. Traditional approaches address these issues through a two-step process of imputation followed by prediction. We propose MAGIC (**M**ulti-t**A**sk **G**aussian Process for **I**mputation and **C**lassification), a novel unified framework that simultaneously performs class-informed missing value imputation and label prediction within a hierarchical multi-task Gaussian process coupled with functional logistic regression. To handle intractable likelihood components, MAGIC employs Taylor expansion approximations with bounded error analysis, and parameter estimation is performed using EM algorithm with block coordinate optimization supported by convergence analysis. We validate MAGIC through two healthcare applications: prediction of post-traumatic headache improvement following mild traumatic brain injury and prediction of in-hospital mortality within 48 hours after ICU admission. In both applications, MAGIC achieves superior predictive accuracy compared to existing methods. The ability to generate real-time and accurate predictions with limited samples facilitates early clinical assessment and treatment planning, enabling healthcare providers to make more informed treatment decisions.

## 1. Introduction

The analysis of temporal patterns in healthcare data has emerged as an important tool for improving patient diagnosis and management. Time series data has proven to be valuable, enabling improved quality of care and accurate healthcare predictions across various domains (Kaushik et al., 2020; Piccialli et al., 2021; Aydin, 2022). Despite this potential, healthcare time series presents significant challenges that limit the effectiveness of existing approaches. First, time misalignment occurs when data collection schedules vary across participants or when adherence to monitoring intervals is inconsistent. This results in irregularly spaced observations. Second, data sparsity is common since participants often contribute fewer observations than what would be ideal for predictive modeling.

To illustrate these challenges and motivate our proposed method, we use two healthcare applications as motivating examples: post-traumatic headache (PTH) recovery prediction via telemonitoring and predictive modeling in intensive care units (ICU).

*Motivating example I: PTH recovery prediction via telemonitoring*

Telemonitoring leverages information technology to remotely collect and transmit user-specific health data through audio, video, sensors, and other telecommunications technologies (Meystre, 2005; Pare et al., 2007; Chaudhry et al., 2010). By enabling real-time health status monitoring from patients' homes, telemonitoring facilitates continuous medical observation and management while reducing the burden of in-person clinical visits (Hanley et al., 2013; Raso et al., 2021). Our first application examines PTH recovery prediction following mild traumatic brain injury (Ashina et al., 2021; Schwedt, 2021). Recent research has identified altered speech patterns in PTH participants compared to healthy controls, suggesting speech characteristics as potential predictors of recovery outcomes (Chong et al., 2021). We examine speech patterns, specifically speaking rate, and headache intensity over a six-week period to predict PTH improvement within three months. Although our structured protocols required speech tasks every three days, participants did not provide recordings for more than 70% of scheduled time points.

*Motivating example II: ICU mortality prediction*

Predictive modeling in ICU settings serves multiple critical functions, including mortality risk assessment, post-traumatic stress disorder screening, bacteremic sepsis identification, and length of stay estimation (Kar et al., 2021; Kauppi et al., 2016; Papini et al., 2018; Rahman et al., 2020). Early and accurate predictions enable healthcare providers to implement more effective management strategies and deliver targeted patient care (Kishore et al., 2023). Our second application addresses ICU mortality prediction using the PhysioNet Challenge 2012 dataset, which includes time series features such as vital signs and laboratory values sampled at various intervals. While the primary objective is to predict in-hospital mortality within 48 hours after admission, extensive missing data presents significant modeling challenges. Due to staff availability and hospital protocols, irregular sampling patterns complicate both data preprocessing and predictive modeling.

Various approaches have been used to address the challenges of time misalignment and data sparsity. Interpolation methods are straightforward to implement but rely only on local information, failing to capture broader temporal dynamics (Banerjee & Gelfand, 2002; Moritz et al., 2015). Autoregressive models improve upon interpolation by modeling temporal dependences across observations, but they continue to struggle with irregularly spaced data and sparse observations (Bashir & Wei, 2018; Zhang et al., 2022). There are some deep learning-based approaches that can accommodate irregular time intervals (Weerakody et al., 2021; Kazijevs & Samad, 2023). However, their performance can depend on the availability of sufficiently large training datasets, which may be limited in healthcare applications. In contrast, Gaussian processes (GP) are well suited for capturing structured temporal dependencies, accounting for time misalignment, and performing well with sparse data. Prior GP-based classification and latent-process models have demonstrated that latent trajectories and class labels can be modeled jointly (S. C.-X. Li & Marlin, 2016; Futoma et al., 2017). However, they do not employ the supervised hierarchical MTGP formulation to allow class-label information to guide the reconstruction of irregular time series. To address this gap, this paper proposes **M**ulti-t**A**sk **G**aussian Process for **I**mputation and **C**lassification (MAGIC), a supervised hierarchical MTGP framework that jointly performs missing value imputation and class label prediction. This unified estimation procedure allows imputation and classification to inform each other rather than treating them as independent stages. MAGIC aims to accurately predict missing values by including class-specific information while simultaneously predicting class labels from the imputed time series. The key contributions of this paper are summarized as follows:

*Innovative integrated framework:* MAGIC integrates two tasks, time series imputation and classification, into a unified framework. MAGIC establishes a reciprocal relationship where class label information improves imputation quality while imputed time series enhances classification accuracy. Beyond the integrated design, MAGIC introduces hierarchical Multi-task Gaussian Processes (MTGP) formulation combined with functional logistic regression. The intractable label likelihood component in the likelihood function was addressed using Taylor expansion approximations, and an EM algorithm with block coordinate optimization scheme was proposed for parameter optimization. Furthermore, MAGIC provides theoretical guarantees including approximation error boundedness and algorithm convergence analyses. This supervised hierarchical GP framework uses class labels to guide trajectory imputation, with all model parameters jointly estimated.

*Early prediction capabilities:* MAGIC's ability to handle missing values allows for accurate predictions even when only limited data is observed at early time points. This capability is particularly valuable in healthcare applications, where early predictions can significantly influence patient care decisions and treatment outcomes. The model's predictive performance continues to improve, making it suitable for real-time monitoring applications.

*Clinical impact and practical application*: MAGIC demonstrates superior performance in two real-world clinical applications with incomplete data, time misalignment, and limited sample sizes: (i) PTH recovery prediction and (ii) ICU mortality prediction. Across both tasks, MAGIC consistently outperforms the competing imputation and classification methods even under severe missingness. This enhanced performance enables healthcare providers to make more informed treatment decisions in healthcare environments.

The remainder of this paper is organized as follows: Section 2 reviews related works on GP and functional regression. Section 3 presents preliminaries for GP and functional logistic regression. Section 4 introduces the development of the MAGIC model. Section 5 provides a simulation study. Section 6 illustrates case studies. Section 7 concludes the paper.

## 2. Related works

### 2.1. *Gaussian process*

GP represent a powerful statistical tool for time series analysis, utilizing mean and covariance functions to capture similarities between observations (Williams & Rasmussen, 1995). Single GP (SGP) can predict missing values by conditioning the joint Gaussian prior distribution on observed data. Beyond prediction, GP also serves as a Bayesian non-parametric framework for time series modeling, allowing domain knowledge to be included through kernel and mean function design (Roberts et al., 2013). However, it struggles with extrapolating beyond observed data and requires a separate classification step for class label prediction. To overcome this limitation, the spectral mixture kernel has been proposed, which leverages the Fourier transform of the kernel to model the spectral density of the data with a Gaussian mixture, enabling GP to extrapolate beyond the training horizon (Wilson & Adams, 2013).

From another perspective, MTGP enhance this framework by modeling shared covariance matrices to capture inter-task dependencies (Bonilla et al., 2007; Williams et al., 2008). Later innovations included self-measuring similarity in covariance functions and introduced latent variables with Expectation-Maximization (EM)-like estimation to address data sparsity (Hayashi et al., 2012), and subsequent applications extended this framework to multi-trait, multi-environment imputation tasks (Hori et al., 2016). A nonparametric Bayesian causal inference method within MTGP utilizes factual and counterfactual outcomes in treatment settings, employing risk-based Empirical Bayes to adapt the prior for joint error minimization (Alaa & Van Der Schaar, 2017). More recently, MTGP with common mean employs a shared mean process across samples and the EM algorithm for estimation of common and individual parameters, reducing parameter estimation complexity through common mean process (Leroy et al., 2022). However, these approaches still do not include class labels during imputation, creating a gap between imputation and prediction tasks. An extension of this framework introduced cluster-specific mean processes, where tasks are probabilistically assigned to latent cluster through a variational EM algorithm (Leroy et al., 2023). While this approach integrates clustering and prediction within a unified MTGP framework, its latent clusters are not observed outcome labels and therefore do not define label-specific population mean processes. Other studies have directly coupled GP-based representations with classifiers. An uncertainty-aware GP adapter has been jointly trained with a downstream classifier (S. C.-X. Li & Marlin, 2016). An MTGP has also been combined with an RNN classifier for clinical prediction (Futoma et al., 2017). These studies demonstrate that GP-based modeling and classification need not always be performed as separate stages, but they do not use observed class labels to construct class-specific population mean processes for imputation. This distinction motivates a framework that explicitly incorporates supervised class information into hierarchical trajectory estimation.

### *2.2. Functional regression*

Functional regression provides a natural choice, as it extends traditional regression to analyze relationships between predictors and responses in a continuous domain (Morris, 2015). For binary outcome prediction, functional predictor logistic regression employs truncated basis function expansion to reduce predictor dimensionality (Ramsay & Silverman, 2006). A methodological advancement addressed multicollinearity issues for highly correlated covariates (Escabias et al., 2004). To further address multicollinearity issues, the functional partial least squares logistic regression model has been proposed as an alternative to functional principal component approaches, providing improved parameter estimation (Escabias et al., 2007). The inclusion of penalized terms in these regression methods has improved their capability to address sparsity and smoothness challenges (Harezlak et al., 2007). Penalized spline approaches for functional logit regression were proposed to integrate smoothed function principal component (FPC) analysis with penalized likelihood estimation (Aguilera-Morillo et al., 2013). Additionally, fused lasso penalty was included in the functional logistic regression to simultaneously perform classification and select informative curve intervals (H. Kim & Kim, 2018).

Despite these advantages, these models show limitations in handling sparse or irregularly spaced data. The functional principal components analysis through conditional expectation method provides a framework for deriving covariance functions and estimating covariance surfaces based on observed data (Yao et al., 2005). This method utilizes conditional expectation to compute FPC scores and predict trajectories. However, this unsupervised approach does not consider the response variable. To address this, robust principal component functional logistic regression was introduced to integrate principal component extraction with logistic regression in a supervised manner while enhancing robustness against outliers (Denhere & Billor, 2016). Nonetheless, this approach does not account for sparse or irregularly observed functional data. A supervised sparse extension was developed to handle missing values while including supervision information and penalty functions (G. Li et al., 2016). However, this approach prioritizes FPC extraction over prediction accuracy optimization.

### *2.3. Gaps in existing research*

Current methods in both MTGP and functional logistic regression exhibit substantial limitations. MTGP frameworks do not leverage class label information during imputation and require supplementary classification algorithms. Similarly, functional logistic regression models struggle to directly address missing values, thus requiring additional imputation steps. These limitations have led to two-step approaches: SGP with functional logistic regression and MTGP with functional logistic regression. While these approaches manage misaligned timestamps and sparse data, they neither utilize label information during imputation nor optimize parameters jointly across the combined models. Label-agnostic imputation may consequently attenuate class-discriminative temporal patterns, particularly when they occur in sparsely observed regions. However, joint modeling is also non-trivial because reconstruction and classification optimize different objectives, and improving classification performance does not necessarily produce meaningful imputations (S. Kim et al., 2023). In contrast, our proposed MAGIC framework incorporates class label information into the imputation process through class-specific population mean functions and jointly optimizes the imputation and classification parameters. MAGIC therefore avoids a separate two-step procedure while improving predictive accuracy.

## 3. Preliminaries

### *3.1. Gaussian process*

Consider a time series $y(t)$ (or equivalently, $y$) observed at time points $t = \{t_1, \dots, t_n\} \subset T$, where $T$ denotes the global time domain. Define the unobserved time points $t^* = T \setminus t = \{t_1^*, \dots, t_{n^*}^*\}$ such that $t \cap t^* = \emptyset$. Let $y(t) = [y(t_1), \cdots, y(t_n)] \in \mathbb{R}^n$. Under a GP assumption, $y(t)$ follows a multivariate normal distribution:

$$y(t) \sim N\left(m(t), K_\theta^t\right), \tag{1}$$

where $m(t) = [m(t_1), \cdots, m(t_n)]$ is the mean vector, and $K_\theta^t \in \mathbb{R}^{n \times n}$ is the covariance matrix. Each element of $K_\theta^t$ is computed using the Radial Basis Function (RBF) kernel:

$$k(t,t') = \theta_v^2 \exp(-\frac{(t-t')^2}{2\theta_l^2}), \tag{2}$$

where $\theta = \{\theta_v, \theta_l\}$ are the hyperparameters. $K_\theta^t$ forms a symmetric covariance matrix that characterizes the dependencies between values at different time points. The amplitude $\theta_v$ controls the overall scale of function variations, and the length-scale $\theta_l$ determines how rapidly correlations decay as the time difference increases. The RBF kernel is used as the default because it provides smooth temporal interpolation between sparse observations. MAGIC can also accommodate alternative covariance kernels, and Section 5.3.2 evaluates its sensitivity using the Matérn-5/2 kernel.

To impute missing values at unobserved time points, the joint prior distribution is defined as:

$$\begin{pmatrix} y(t) \\ y(t^*) \end{pmatrix} \sim N\left(\begin{pmatrix} m(t) \\ m(t^*) \end{pmatrix} \begin{pmatrix} K_\theta^{(t,t)} + \sigma^2 I & K_\theta^{(t,t^*)} \\ K_\theta^{(t^*,t)} & K_\theta^{(t^*,t^*)} \end{pmatrix}\right). \tag{3}$$

The posterior mean and covariance of $y(t^*)$ are given by:

$$\begin{gathered} \mathbb{E}[y(t^*)] = m(t^*) + K_\theta^{(t^*,t)} \left[K_\theta^{(t,t)} + \sigma^2 I\right]^{-1} \left(y(t) - m(t)\right), \\ Cov\left(y(t^*)\right) = K_\theta^{(t^*,t^*)} - K_\theta^{(t^*,t)} \left[K_\theta^{(t,t)} + \sigma^2 I\right]^{-1} K_\theta^{(t,t^*)}. \end{gathered} \tag{4}$$

These equations provide a posterior estimate of $y(t^*)$ by conditioning on observed values, allowing for function interpolation at any unobserved time point.

### *3.2. Functional logistic regression*

Let $K$ be the number of basis functions. Define the basis vector $\phi(t) = [\phi_1(t), \cdots, \phi_K(t)]^T \in \mathbb{R}^K$ and the functional coefficient vector $\beta_1 = [\beta_{11}, \cdots \beta_{1K}]^T \in \mathbb{R}^K$. We represent the time-varying coefficient:

$$\beta_1(t) = \textstyle\sum_{k=1}^K \beta_{1k}\phi_k(t) = \phi(t)^T \beta_1. \tag{5}$$

For sample $i$, let $y_i(t)$ denote the time series predictor and $z_i \in \{0, 1\}$ be the corresponding binary class label. Define the predictor $x_i = \left[1 \quad \int_T \phi(t) y_i(t) dt\right]^T \in \mathbb{R}^{K+1}$. The probability of class $z_i = 1$ is modeled using the logistic function, where the log-odds are given by:

$$\begin{aligned} \log\left(\frac{p(z_i = 1|y_i)}{1 - p(z_i = 1|y_i)}\right) &= \beta_0 + \beta_1^T \int_T \phi(t) y_i(t) dt \\ &= x_i^T \beta, \end{aligned} \tag{6}$$

where $\beta_0$ is an intercept term, $\beta = [\beta_0 \quad \beta_1^T]^T$ is the coefficient vector. Hence, the probability of $z_i = 1$ is then:

$$p(z_i = 1|y_i) = \frac{1}{1 + \exp(-x_i^T \beta)}. \tag{7}$$

The corresponding log-likelihood function is given by:

$$\begin{aligned} l(\beta) &= \log\left(\prod_i p(z_i = 1|y_i)^{z_i} \cdot \left(1 - p(z_i = 1|y_i)\right)^{1-z_i}\right) \\ &= \sum_i z_i x_i^T \beta - \log(1 + \exp(x_i^T \beta)). \end{aligned} \tag{8}$$

Maximizing $l(\beta)$ yields the maximum likelihood estimates for the parameter vector β, which can be used for classification.

## 4. Development of MAGIC model

Let $i$ be the index for individuals, $i = 1,2,\cdots,N$. For each individual $i$, we observe a time series $y_i(t) = [y_i(t_{i1}),\cdots,y_i(t_{in_i})]$ (or equivalently, $y$) at time points $t_i = \{t_{i1},\cdots,t_{in_i}\} \subset T$, where $t_i$ represents the set of observed time points for individual $i$, and $T$ denotes the global time domain. The set of missing or unobserved time points for individual $i$ is denoted as $t_i^* = T \setminus t_i$ such that $t \cap t^* = \emptyset$. Due to variability in data collection, individuals may have misaligned time points, meaning that $t_i$ and $t_j$ can differ across individuals. Each individual belongs to one of two classes, denoted by a binary outcome $z_i \in \{0,1\}$. Let $Y = \{y_1(t), y_2(t),\cdots,y_N(t)\}$ and $Z = \{z_1, z_2,\cdots,z_N\}$ be the overall collection of time series and the corresponding binary outcomes. To distinguish between classes, define $Y_0 = \{y_i(t)|z_i = 0\}$ and $Y_1 = \{y_i(t)|z_i = 1\}$, which are the subsets of $Y$ containing all individuals in class 0 and class 1, respectively. Let $n_0 = |Y_0|$ and $n_1 = |Y_1|$ denote the number of individuals in class 0 and class 1, respectively, with $n_0 + n_1 = N$.

### *4.1.Mathematical formulation*

We propose a hierarchical GP framework, where each individual's time series is decomposed into three components: a class-specific term, an individual-specific term, and a noise term:

$$y_i(t) = \mu_{z_i}(t) + \delta_i(t) + \epsilon_i(t). \tag{9}$$

The term $\mu_{z_i}(t)$ (or equivalently, $\mu_{z_i}$) represents the class-specific GP prior, modeling the shared mean structure within each class. Specifically, for each class:

$$\mu_{z_i}(t) \sim GP\left(m_{z_i}(t), K_{\theta_{z_i}}^t\right), \tag{10}$$

where $m_{z_i}(t)$ (or equivalently, $m_{z_i}$) is the class-specific mean process, and $K^t_{\theta_{z_i}}$ is the covariance kernel matrix that captures temporal dependencies within each class. $\delta_i(t) \sim GP(0, K^t_\theta)$ represents the individual-specific process, where the covariance kernel matrix $K^t_\theta$ accounts for individual temporal variability. The noise term $\epsilon_i(t) \sim N(0, \sigma^2 I)$ represents independent Gaussian noise.

Assume that $\delta_i(t)$ are independent across individuals, and $\epsilon_i(t)$ are also independent across individuals, and $\mu_0(t)$, $\mu_1(t)$, $\delta_i(t)$, and $\epsilon_i(t)$ are mutually independent. Each observed time series, conditioned on the class mean process, follows a Gaussian distribution:

$$y_i(t) | \mu_{z_i}(t) \sim N\left(\mu_{z_i}(t), K^t_\theta + \sigma^2 I\right). \tag{11}$$

The overall likelihood can be written as follows:

$$\begin{aligned} L(\Theta; \mu_0, \mu_1, Y, Z) &= P(Y, Z | \mu_0, \mu_1, \Theta) \cdot P(\mu_0, \mu_1 | \Theta) \\ &= P(Z | Y, \mu_0, \mu_1, \Theta) \cdot P(Y | \mu_0, \mu_1, \Theta) \cdot P(\mu_0, \mu_1 | \Theta) \\ &= \prod_i \left[ P\left(z_i \middle| y_i, \mu_{z_i}, \Theta\right)^{z_i} \cdot \left(1 - P\left(z_i \middle| y_i, \mu_{z_i}, \Theta\right)\right)^{1-z_i} \cdot P\left(y_i \middle| \mu_{z_i}, \Theta\right) \right] \cdot P(\mu_0 | \Theta) \\ &\cdot P(\mu_1 | \Theta), \end{aligned} \tag{12}$$

where $\mu_0$ and $\mu_1$ are unobserved latent mean functions, and $\Theta = \{\theta_0, \theta_1, \theta, \sigma^2\}$ is the set of unknown hyperparameters. Each parameter consists of the RBF kernel parameters $\theta_0 = (\theta_{0,v}, \theta_{0,l})$, $\theta_1 = (\theta_{1,v}, \theta_{1,l})$, $\theta = (\theta_v, \theta_l)$. To delineate the likelihood function, we require four component distributions: $P\left(z_i \middle| y_i, \mu_{z_i}, \Theta\right)$, $P\left(y_i \middle| \mu_{z_i}, \Theta\right)$, $P(\mu_0 | \Theta)$, and $P(\mu_1 | \Theta)$. The latter three probabilities have already been specified in (10) and (11). The remaining task is to characterize the first probability, $P\left(z_i \middle| y_i, \mu_{z_i}, \Theta\right)$, which relates the binary label to the individual time series.

As the observed time series $y_i$ typically contains missing values and the binary-functional relationship naturally suggests a functional logistic regression framework, we construct the complete-data log-likelihood function through a three-step approach: (1) imputing missing values to obtain complete time series trajectories, (2) applying functional logistic regression with the complete time series predictors and the binary outcomes, and (3) assembling all components into the complete-data log-likelihood.

Missing value imputation

We initially impute missing values to obtain a complete time series. The joint distribution of observed and unobserved time points for subject $i$ and corresponding conditional (posterior) mean are given by:

$$\begin{pmatrix} y_i(t_i) \\ y_i(t_i^*) \end{pmatrix} | \mu_{z_i}, \Theta \sim N\left(\begin{pmatrix} \mu_{z_i}(t_i) \\ \mu_{z_i}(t_i^*) \end{pmatrix} \begin{pmatrix} K_\theta^{(t_i,t_i)} + \sigma^2 I & K_\theta^{(t_i,t_i^*)} \\ K_\theta^{(t_i^*,t_i)} & K_\theta^{(t_i^*,t_i^*)} \end{pmatrix}\right) \tag{13}$$

$$\mathbb{E}\left[y_i(t_i^*) \middle| y_i, \mu_{z_i}, \Theta\right] = \mu_{z_i}(t_i^*) + K_\theta^{(t_i^*,t_i)} \left(K_\theta^{(t_i,t_i)} + \sigma^2 I\right)^{-1} \left(y_i(t_i) - \mu_{z_i}(t_i)\right) \tag{14}$$

To obtain a complete time series including observed and imputed values, we define:

$$f_i(t) \equiv \begin{cases} y_i(t), & if\ t \in t_i \\ \mathbb{E}\left[y_i(t) \middle| y_i, \mu_{z_i}, \Theta\right], & if\ t \in t_i^* \end{cases} \tag{15}$$

Equation (15) uses the conditional posterior mean as a plug-in estimate for subject-specific missing values. Consequently, the functional logistic regression does not integrate over the conditional residual distribution of these missing values.

*Functional logistic regression*

We then model $p\left(z_i | y_i, \mu_{z_i}, \Theta\right)$ using functional logistic regression with complete time series $f_i(t)$ available. Similar to Section 2.2, let the basis vector be $\phi(t) \in \mathbb{R}^K$ and the coefficient vector $\beta_1 \in \mathbb{R}^K$. Writing in coefficient-first form $\beta_1(t) = \beta_1^T \phi(t)$, the log-odds are:

$$\begin{aligned} \log\left(\frac{p\left(z_i | y_i, \mu_{z_i}, \Theta\right)}{1 - p\left(z_i = 1 \middle| y_i, \mu_{z_i}, \Theta\right)}\right) &= \beta_0 + \beta_1^T \int_T \phi(t) f_i(t) dt \\ &= x_i^T \beta, \end{aligned} \tag{16}$$

where $\beta_0$ is an intercept term, $\beta = [\beta_0 \quad \beta_1^T]^T$ is the coefficient vector, $x_i = \left[1 \quad \int_T \phi(t) f_i(t) dt\right]^T$, $\Theta = \{\theta_0, \theta_1, \theta, \sigma^2, \beta\}$ is the full parameter set. Thus, the log-likelihood for the probability, $P\left(z_i \middle| y_i, \mu_{z_i}, \Theta\right)$, under the functional logistic regression model is given by:

$$\log\left(\prod_i P\left(z_i \middle| y_i, \mu_{z_i}, \Theta\right)^{z_i} \cdot \left(1 - P\left(z_i \middle| y_i, \mu_{z_i}, \Theta\right)\right)^{1-z_i}\right) = \sum_i z_i x_i^T \beta - \log(1 + \exp(x_i^T \beta)). \tag{17}$$

*Complete-data log-likelihood function*

We construct the complete-data log-likelihood function $l(\Theta)$ from (12) using established components, which includes the label likelihood term from (17), the individual time series likelihood term from (11) and the class mean prior terms from (10):

$$
\begin{aligned}
l(\Theta) = &\sum_i \left[ \overbrace{z_i x_i^T \beta - \log\left(1 + \exp\left(x_i^T \beta\right)\right)}^{Label\ likelihood} \right] \\
&- \frac{1}{2} \sum_i \left[ \overbrace{\log(|K_\theta + \sigma^2 I|) + \left(y_i - \mu_{z_i}\right)^T (K_\theta + \sigma^2 I)^{-1} \left(y_i - \mu_{z_i}\right)}^{Complete\ time\ series\ likelihood} \right] \\
&- \frac{1}{2} \left[ \overbrace{\log\left(\left|K_{\theta_0}\right|\right) + (\mu_0 - m_0)^T K_{\theta_0}^{-1} (\mu_0 - m_0)}^{Class\ 0\ mean\ prior} \right] \\
&- \frac{1}{2} \left[ \overbrace{\log\left(\left|K_{\theta_1}\right|\right) + (\mu_1 - m_1)^T K_{\theta_1}^{-1} (\mu_1 - m_1)}^{class\ 1\ mean\ prior} \right] - \frac{(N+2)k}{2} \log(2\pi).
\end{aligned}
\tag{18}
$$

### *4.2. Estimation*

As $l(\Theta)$ includes the latent variables $\mu_0$ and $\mu_1$, we employ the Expectation-Maximization (EM) algorithm, an iterative approach for parameter optimization (Wu, 1983). In the Expectation step (E-step), we compute the expected value of the complete-data log-likelihood function conditioned on the observed data and the current parameter estimates. In the Maximization step (M-step), we optimize the expected complete-data log-likelihood calculated in the E-step with respect to the parameters to update their estimates. This iterative process between E-step and M-step continues until convergence is achieved.

#### *4.2.1. E-step*

Given that the hyperparameters are initialized or have been estimated from a previous M-step, we define the expectation of the complete-data log-likelihood from (18) with respect to the latent class-level mean functions, $\mu_0$ and $\mu_1$:

$$
Q\left(\Theta \middle| \Theta^{(r-1)}\right) = \mathbb{E}_{\mu_0, \mu_1 | Y, Z, \Theta^{(r-1)}} [l(\Theta)] \tag{19}
$$

When working with limited sample sizes and high missing ratios, the class-level mean functions can produce highly fluctuating curves that overfit sparse observations. To mitigate such fluctuations, we introduce a smoothing penalty that regularizes $\mu_0(t)$ and $\mu_1(t)$. We add terms $\frac{1}{2}\mu_0^T R \mu_0$ and $\frac{1}{2}\mu_1^T R \mu_1$ to the negative log-posterior.

Let $D_2$ denote the unscaled second-difference matrix. We define the difference operator as $D(\rho) = D_2/\rho^2$, and the corresponding penalty matrix as $R(\rho) = \rho^{-4} D_2^T D_2$, where $\rho > 0$ is a smoothing-scale parameter. Therefore, smaller values of $\rho$ impose stronger smoothing, whereas larger values produce weaker regularization. This penalty approximates the continuous-time roughness penalty $\int \left(\mu''(t)\right)^2 dt$, encouraging temporal smoothness in the estimated class-level mean curves. Please see Appendix A for the definition and structure of $R$. Therefore, the distributions of the class-level mean functions used in the E-step at iteration $r$ are summarized in Proposition 1. Please see the proof in Appendix A.

**Proposition 1.** At the $r$-th iteration of the EM algorithm, the posterior distributions of the class-level mean functions $\mu_0$ and $\mu_1$ are Gaussian:

$$\begin{aligned} p\left(\mu_0 \middle| Y_0, Z_0, \Theta^{(r-1)}\right) &= N\left(\tilde{m}_0, \tilde{K}_0\right), \\ p\left(\mu_1 \middle| Y_1, Z_1, \Theta^{(r-1)}\right) &= N\left(\tilde{m}_1, \tilde{K}_1\right), \end{aligned} \tag{20}$$

where

$$\begin{aligned} \tilde{m}_0 &= \tilde{K}_0 \left( K^{-1}_{\theta_0^{(r-1)}} m_0 + \sum_{i:z_i=0} \left(K_{\theta^{(r-1)}} + \left(\sigma^{(r-1)}\right)^2 I\right)^{-1} y_i \right), \\ \tilde{K}_0 &= \left( K^{-1}_{\theta_0^{(r-1)}} + R + n_0 \left(K_{\theta^{(r-1)}} + \left(\sigma^{(r-1)}\right)^2 I\right)^{-1} \right)^{-1}, \\ \tilde{m}_1 &= \tilde{K}_1 \left( K^{-1}_{\theta_1^{(r-1)}} m_1 + \sum_{i:z_i=1} \left(K_{\theta^{(r-1)}} + \left(\sigma^{(r-1)}\right)^2 I\right)^{-1} y_i \right), \\ \tilde{K}_1 &= \left( K^{-1}_{\theta_1^{(r-1)}} + R + n_1 \left(K_{\theta^{(r-1)}} + \left(\sigma^{(r-1)}\right)^2 I\right)^{-1} \right)^{-1}. \end{aligned} \tag{21}$$

#### 4.2.2. M-step

Given the posterior mean and covariance of each class-level mean function computed in E-step at iteration $r$, we proceed in the M-step by jointly estimating the class-level GP parameters $\theta_0$ and $\theta_1$, the individual-level GP parameters $\theta$ and $\sigma^2$, and the functional logistic regression coefficient $\beta$ under the common Q-function defined in (19). Joint estimation refers to coupling the GP and classification parameters under a common expected complete-data objective, while optimizing them through coordinated block updates. This leads to the following optimization problem with a regularization penalty on $\beta_1$:

$$\Theta^{(r)} = arg \max_{\Theta} Q\left(\Theta \middle| \Theta^{(r-1)}\right) - \frac{\lambda}{2} \|\beta_1\|_2^2 \tag{22}$$

The M-step involves three key components: (1) Taylor expansion approximation to handle the intractable expectation in the label likelihood component, (2) boundedness analysis of the approximation error and regularization, and (3) block coordinate optimization to decompose the high-dimensional optimization problem into four manageable subproblems. We detail each component below.

<u>*Taylor expansion approximation*</u>

The main challenge in solving (22) arises from the label likelihood component in (18) and (19), $\mathbb{E}[z_i x_i^T \beta] - \mathbb{E}[\log(1 + \exp(x_i^T \beta))]$. While we can directly compute $\mathbb{E}[z_i x_i^T \beta]$ due to the linearity, the term $\mathbb{E}[\log(1 + \exp(x_i^T \beta))]$ involves the expectation of a nonlinear function of $x_i^T \beta$, which cannot be computed analytically. To address this computational challenge, we employ a Taylor expansion around $\mathbb{E}[x_i^T \beta]$ to approximate this intractable expectation.

We first need to derive both the mean and variance of $x_i^T \beta$. The mean $U_i = \mathbb{E}[x_i^T \beta]$ is required for computing the first term $\mathbb{E}[z_i x_i^T \beta]$ in the label likelihood component. The variance $V_i = Var[x_i^T \beta]$ is essential for the Taylor expansion approximation of $\mathbb{E}[\log(1 + \exp(x_i^T \beta))]$, as higher-order terms in the expansion depend on the variance $V_i$. Proposition 2 summarizes these moment calculations and the resulting approximation for the label likelihood component. Please see Appendix B for the proof.

**Proposition 2.** Let $U_i = \mathbb{E}[x_i^T \beta]$ and $V_i = Var[x_i^T \beta]$ denote the mean and variance of the linear predictor $x_i^T \beta$, which are given by:

$$U_i = \beta_0 + \beta_1^T \int_T \phi(t)[\tilde{m}_{z_i}(t) + K_\theta^{(t^*,t_i)} \left(K_\theta^{(t_i,t_i)} + \sigma^2 I\right)^{-1} \left(y_i - \tilde{m}_{z_i}(t_i)\right)]dt \tag{23}$$

$$V_i = \beta_1^T \left( \int_T \int_T \phi(t)\phi(t')^T \left(\tilde{K}_{z_i} - B_i \tilde{K}_{z_i} - \tilde{K}_{z_i} B_i^T + B_i \tilde{K}_{z_i} B_i^T\right) dt dt' \right) \beta_1, \tag{24}$$

where $B_i = K_\theta^{(t^*,t_i)} \left(K_\theta^{(t_i,t_i)} + \sigma^2 I\right)^{-1}$ and $\tilde{K}_{z_i} = Cov[\mu_{z_i}(t), \mu_{z_i}(t')]$

The expectation $\mathbb{E}[\log(1 + \exp(x_i^T \beta))]$ has the following second-order Taylor approximation:

$$\begin{aligned} &\mathbb{E}[\log(1 + \exp(x_i^T \beta))] \\ &\qquad \approx \log\left(1 + \exp(U_i) + \frac{1}{2}\exp(U_i)\, V_i\right) - \frac{\exp(2U_i) V_i}{2\left(1 + \exp(U_i) + \frac{1}{2}\exp(U_i)\, V_i\right)^2} \end{aligned} \tag{25}$$

The variance $V_i$ propagates posterior uncertainty in the latent class-level mean functions into the expected label likelihood. However, it does not include the conditional residual variance of the subject-specific unobserved values, which are replaced by their posterior means in Equation (15).

<u>*Boundedness analysis and regularization*</u>

As the Taylor expansion introduces remainder terms, it is necessary to establish their boundedness to ensure approximation quality and motivate appropriate regularization. The following propositions provide this theoretical foundation.

**Proposition 3.** Consider the complete time series $f_i(t)$ defined in (15), where missing values are imputed using GP posterior means. Since $f_i(t)$ is derived from GP, the linear predictor $x_i^T\beta$ defined in (16) follows a sub-Gaussian distribution. For the second-order Talor expansion of $\mathbb{E}\left[\log\left(1+\exp\left(x_i^T\beta\right)\right)\right]$ used in Proposition 2, the remainder terms are bounded by a constant multiple of $\left(Var\left[x_i^T\beta\right]\right)^{3/2}$.

**Proposition 4.** Under the conditions in Proposition 3,

$$Var\left[x_i^T\beta\right] = O(\|\beta_1\|_2^2).$$

Please see the proofs in Appendices C and D. Combining Propositions 3 and 4 demonstrates that the remainder terms in the Taylor expansions are of order $O(\|\beta_1\|_2^3)$, indicating that the approximation accuracy improves as $\beta_1$ decreases in magnitude. Therefore, we include an $l_2$-penalty on $\beta_1$ in the optimization, which ensures that the higher-order terms in the Taylor expansion remain small.

*Block coordinate optimization*

To further expanding the $Q\left(\Theta\middle|\Theta^{(r-1)}\right)$ function, we use the following identity to handle the class mean prior terms, which can also be applied to complete time series likelihood term:

$$\begin{aligned}\mathbb{E}\left[(\mu_0-m_0)^T K_{\theta_0}^{-1}(\mu_0-m_0)\right] &= \mathbb{E}\left[Tr\left(K_{\theta_0}^{-1}(\mu_0-m_0)(\mu_0-m_0)^T\right)\right]\\ &= Tr\left(\widetilde{K}_0 K_{\theta_0}^{-1}\right) + Tr\left(K_{\theta_0}^{-1}(\widetilde{m}_0-m_0)(\widetilde{m}_0-m_0)^T\right)\\ &= Tr\left(\widetilde{K}_0 K_{\theta_0}^{-1}\right) + (\widetilde{m}_0-m_0)^T K_{\theta_0}^{-1}(\widetilde{m}_0-m_0)\end{aligned} \qquad (26)$$

Including Proposition 2 together with (26), we can rewrite the $Q$-function as follows:

$$\begin{aligned}Q\left(\Theta\middle|\Theta^{(r-1)}\right) = &\sum_i \mathcal{L}_i\\ &-\frac{1}{2}\sum_i \left[\log\left(\left|K_{\theta^{(r-1)}} + \left(\sigma^{(r-1)}\right)^2 I\right|\right) + Tr\left(\widetilde{K}_{z_i}\left(K_{\theta^{(r-1)}} + \left(\sigma^{(r-1)}\right)^2 I\right)^{-1}\right)\right.\\ &\left.+\left(y_i-\widetilde{m}_{z_i}\right)^T\left(K_{\theta^{(r-1)}} + \left(\sigma^{(r-1)}\right)^2 I\right)^{-1}\left(y_i-\widetilde{m}_{z_i}\right)\right] - \frac{1}{2}\log\left(\left|K_{\theta_0^{(r-1)}}\right|\right)\\ &-\frac{1}{2}\left[Tr\left(\widetilde{K}_0 K_{\theta_0^{(r-1)}}^{-1}\right) + (\widetilde{m}_0-m_0)^T K_{\theta_0^{(r-1)}}^{-1}(\widetilde{m}_0-m_0)\right] - \frac{1}{2}\log\left(\left|K_{\theta_1^{(r-1)}}\right|\right)\\ &-\frac{1}{2}\left[Tr\left(\widetilde{K}_1 K_{\theta_1^{(r-1)}}^{-1}\right) + (\widetilde{m}_1-m_1)^T K_{\theta_1^{(r-1)}}^{-1}(\widetilde{m}_1-m_1)\right] + C,\end{aligned} \qquad (27)$$

where $\mathcal{L}_i = z_i U_i - \log\left(1+\exp U_i + \frac{1}{2}\exp(U_i)V_i\right) + \frac{\exp(2U_i)V_i}{2\left(1+\exp U_i+\frac{1}{2}\exp(U_i)V_i\right)^2}$ and $C \in \mathbb{R}$ is a constant.

We apply a conditional maximization scheme to perform block coordinate ascent over the parameter blocks. To decompose the $Q$-function in (27), note that $\theta_0$ appears only in the Gaussian log-likelihood term, $-\frac{1}{2}\log\left(\left|K_{\theta_0^{(r-1)}}\right|\right)-\frac{1}{2}Tr\left(\widetilde{K}_0K_{\theta_0^{(r-1)}}^{-1}\right)-\frac{1}{2}(\widetilde{m}_0-m_0)^TK_{\theta_0^{(r-1)}}^{-1}(\widetilde{m}_0-m_0)$, which allows us to maximize this term independently of the other parameters. A similar decomposition applies to $\theta_1$. The remaining parameters $(\theta,\sigma^2,\beta)$ are coupled, but only $\beta$ enters through the $\mathcal{L}_i$ term, whereas $(\theta,\sigma^2)$ appear in both $\mathcal{L}_i$ and complete time series likelihood terms. We first update $\beta$ by maximizing $\sum_i\mathcal{L}_i$ with an $l_2$-penalty while keeping $(\theta,\sigma^2)$ fixed. Then, we update $(\theta,\sigma^2)$ by maximizing the remaining terms while holding $\beta$ fixed. This leads to four subproblems:

$$\theta_0^{(r)}=arg\max_{\theta_0}\left\{-\frac{1}{2}\log\left(\left|K_{\theta_0^{(r-1)}}\right|\right)-\frac{1}{2}Tr\left(\widetilde{K}_0K_{\theta_0^{(r-1)}}^{-1}\right)-\frac{1}{2}(\widetilde{m}_0-m_0)^TK_{\theta_0^{(r-1)}}^{-1}(\widetilde{m}_0-m_0)\right\}, \tag{28}$$

$$\theta_1^{(r)}=arg\max_{\theta_1}\left\{-\frac{1}{2}\log\left(\left|K_{\theta_1^{(r-1)}}\right|\right)-\frac{1}{2}Tr\left(\widetilde{K}_1K_{\theta_1^{(r-1)}}^{-1}\right)-\frac{1}{2}(\widetilde{m}_1-m_1)^TK_{\theta_1^{(r-1)}}^{-1}(\widetilde{m}_1-m_1)\right\}, \tag{29}$$

$$\beta^{(r)}=arg\max_{\beta}\sum_i\mathcal{L}_i-\frac{\lambda}{2}\|\beta_1\|_2^2, \tag{30}$$

$$\begin{aligned}\left(\theta^{(r)},\left(\sigma^{(r)}\right)^2\right)=arg\max_{\theta,\sigma^2}\Bigg\{&\sum_i\mathcal{L}_i-\frac{N}{2}\log\left(\left|K_{\theta^{(r-1)}}+\left(\sigma^{(r-1)}\right)^2I\right|\right)\\&-\frac{1}{2}\sum_i\Bigg[Tr\left(\widetilde{K}_{z_i}\left(K_{\theta^{(r-1)}}+\left(\sigma^{(r-1)}\right)^2I\right)^{-1}\right)\\&+\left(y_i-\widetilde{m}_{z_i}\right)^T\left(K_{\theta^{(r-1)}}+\left(\sigma^{(r-1)}\right)^2I\right)^{-1}\left(y_i-\widetilde{m}_{z_i}\right)\Bigg]\Bigg\},\end{aligned} \tag{31}$$

Where $\mathcal{L}_i=z_iU_i-\log\left(1+\exp U_i+\frac{1}{2}\exp(U_i)V_i\right)+\frac{\exp(2U_i)V_i}{2\left(1+\exp U_i+\frac{1}{2}\exp(U_i)V_i\right)^2}$.

At each step, the block is optimized using the limited memory Broyden-Fletcher-Goldfarb-Shanno algorithm with bound constraints (L-BFGS-B) (Nocedal, 1980; Zhu et al., 1997).

### *4.2.3. Algorithm and Convergence Analysis*

The complete MAGIC optimization procedure is summarized below.

| **Algorithm** for solving the MAGIC optimization |
| --- |
| **Input**: Data $Y$ and $Z$; stopping tolerance $\epsilon$; initial parameters $\Theta^0=\{\theta_0^0,\theta_1^0,\theta^0,(\sigma^0)^2,\beta^0\}$.<br>**Output**: Estimated parameters $\widehat{\Theta}$.<br>1. **Initialize**: Set r ← 0.<br>2. **Repeat**<br>3. **E-step**: Compute $\widetilde{m}_0$, $\widetilde{K}_0$, $\widetilde{m}_1$, $\widetilde{K}_1$ using Proposition 1.<br>4. **M-step**: Update parameters to $\Theta^{(r+1)}$ by solving the subproblems in (28)-(31) sequentially. |

5. **Update iteration index**: r ← r + 1.

6. **Until** $\|\Theta^{(r)} - \Theta^{(r-1)}\| < \epsilon$

**Convergence analysis**: To analyze convergence, we apply Theorem 3 of (Meng & Rubin, 1993), which states that every limit point of the sequence is a stationary point provided that three conditions hold: monotonicity, continuity, and space-filling. For monotonicity, our block coordinate update is accepted only if it increases the Q-function value at each iteration. Otherwise, the previous parameter value is retained, which guarantees that the sequence of objective values is non-decreasing. For continuity, the $Q$-function is continuous in each block since all components (e.g., log-determinant, trace, quadratic forms) are continuous in the model parameters. For space-filling, the block coordinate scheme is space-filling, as each step updates one parameter block while keeping the others fixed. Cycling these steps permits movement in every parameter direction. Since these three conditions are satisfied, Theorem 3 of (Meng & Rubin, 1993) ensures that every limit point of the sequence generated by the MAGIC optimization algorithm is a stationary point.

### *4.3. Prediction*

We formulate a maximum a posteriori (MAP) decision problem. By Bayes' rule, we determine the class $z \in \{0,1\}$ that maximizes the posterior probability $p\left(z \middle| y_{new}(t), Y, \widehat{\Theta}\right)$, where $y_{new}$ is a new sample. Taking logarithms reduces this to summing the class-conditional log-likelihood and the class log-prior:

$$\begin{aligned} z_{new} &= argmax_{z\in\{0,1\}} p\left(z \middle| y_{new}(t), Y, \widehat{\Theta}\right) \\ &= argmax_{z\in\{0,1\}} p\left(y_{new}(t) \middle| Y_z, \widehat{\Theta}\right) \cdot p\left(z \middle| Y, \widehat{\Theta}\right) \\ &= argmax_{z\in\{0,1\}} \left[\underbrace{\log p\left(y_{new}(t) \middle| Y_z, \widehat{\Theta}\right)}_{\substack{class-conditional \\ log-likelihood}} + \underbrace{\log p\left(z \middle| Y, \widehat{\Theta}\right)}_{class\ log-prior}\right]. \end{aligned} \tag{32}$$

**Proposition 5.** The multi-task prior distribution of $y_{new}(t)|Y_z, \widehat{\Theta}$ follows a Gaussian, i.e.

$$p\left(y_{new}(t) \middle| Y_{z_{new}}, \widehat{\Theta}\right) = N\left(\widetilde{m}_{z_{new}}(t), \tilde{\Sigma}_{z_{new}}\right),$$

where $\tilde{\Sigma}_{z_{new}} = \widetilde{K}_{z_{new}} + K_{\widehat{\Theta}} + \sigma^2 I.$

The class-conditional log-likelihood can be calculated using Proposition 5. Please see the proof in Appendix E. Additionally, the class log-prior for $p\left(z \middle| Y, \widehat{\Theta}\right)$ can be computed using the fraction of training samples in each class. This allows us to determine the most probable class $z_{new}$. To perform imputation for unobserved time points $t_{new}^*$, given observed time point $t_{new}$, we rewrite the multi-task prior distribution by Proposition 5:

$$\begin{pmatrix} y_{new}(t_{new}) \\ y_{new}(t_{new}^*) \end{pmatrix} | z_{new}, Y, \widehat{\Theta} \sim N\left( \begin{pmatrix} \widetilde{m}_{z_{new}}(t_{new}) \\ \widetilde{m}_{z_{new}}(t_{new}^*) \end{pmatrix} \begin{pmatrix} \widetilde{\Sigma}_{z_{new}}^{(t_{new}, t_{new})} & \widetilde{\Sigma}_{z_{new}}^{(t_{new}, t_{new}^*)} \\ \widetilde{\Sigma}_{z_{new}}^{(t_{new}^*, t_{new})} & \widetilde{\Sigma}_{z_{new}}^{(t_{new}^*, t_{new}^*)} \end{pmatrix} \right),$$
$$\mathbb{E}\left[y_{new}(t) \middle| Y_{z_{new}}, \widehat{\Theta}\right] = \widetilde{m}_{z_{new}}(t) + \widetilde{\Sigma}_{z_{new}}^{(t_{new}^*, t_{new})} \left(\widetilde{\Sigma}_{z_{new}}^{(t_{new}, t_{new})}\right)^{-1} \left(y_{new}(t_{new}) - \widetilde{m}_{z_{new}}(t_{new})\right). \quad (33)$$

Thus, the complete time series for the new sample is constructed as:

$$f_{new}(t) = \begin{cases} y_{new}(t), & if\ t \in t_{new} \\ \mathbb{E}\left[y_{new}(t) \middle| Y_{z_{new}}, \widehat{\Theta}\right], & if\ t \in t_{new}^* \end{cases} \quad (34)$$

Finally, these imputed values can be embedded into the model for class probabilities:

$$\log\left(\frac{p\left(z_{new} | y_{new}, \mu_{z_{new}}, \widehat{\Theta}\right)}{1 - p\left(z_{new} | y_{new}, \mu_{z_{new}}, \widehat{\Theta}\right)}\right) = \hat{\beta}_0 + \hat{\beta}_1^T \int_T \phi(t) f_{new}(t) dt,$$
$$p\left(z_{new} | y_{new}, \mu_{z_{new}}, \widehat{\Theta}\right) = \frac{1}{1 + \exp(-\hat{\beta}_0 - \hat{\beta}_1^T \int_T \phi(t) f_{new}(t) dt)}. \quad (35)$$

This formulation allows us to simultaneously perform missing value imputation and compute predicted probabilities in a unified framework.

## 5. Simulation study

To evaluate MAGIC's performance, we compared it with GP-based, deep learning, and feature-based alternatives. The GP-based methods included SGP and MTGP with a common mean (Leroy et al., 2022), each followed by functional logistic regression using the same cubic B-spline basis functions and $l_2$-penalty terms as those employed in MAGIC. The deep learning methods included GRU-D, a recurrent model with trainable decay mechanisms for missing values (Che et al., 2018), ODE-RNN, a continuous-time recurrent model based on neural ordinary differential equations (Rubanova et al., 2019), STraTS, a Transformer-based model for sparse and irregularly sampled time series (Tipirneni & Reddy, 2022). We also included NRTSI, a non-recurrent attention-based imputation model for irregularly sampled time series (Shan et al., 2023). NRTSI-imputed trajectories were classified using functional logistic regression. We also considered NRTSI-C22, which applied logistic regression to catch22 features extracted from the imputed trajectories (Lubba et al., 2019). Finally, MAGIC-NC replaced the class-specific mean processes with a shared class-independent mean process to evaluate the contribution of class information during imputation. All methods were evaluated using identical data partitions and missingness settings. All computational analyses were executed using Python 3.10.10 on the PACE high-performance computing cluster, with each job allocated four CPU cores, 32GB of system memory, and one NVIDIA Tesla V100 GPU.

### *5.1. Simulation setup*

We considered two data-generating scenarios over 51 equally spaced time points, $t = 0, \cdots, 50$. In both scenarios, class-specific mean trajectories were independently drawn from GPs centered on opposing sinusoidal signals, $\pm a * \sin(\frac{2\pi}{(50/6)} t)$, followed by 75 individual trajectories per class. The first scenario used $a = 0.5$, RBF kernels, and evenly distributed missingness generated by randomly selecting observations within equally spaced temporal bins. The second used $a = 0.3$, Matérn-5/2 kernels, and baseline-dependent Markov missingness.

For the Markov setting, the baseline measurement at $t = 0$ was always retained. Each subject's missingness probability was determined by a logistic function of the standardized baseline value and a subject-specific random effect, with coefficients of 0.5 and a random-effect standard deviation of 0.7. The intercept was calibrated so that the mean subject-level probability matched the target missing ratio. Missingness over the remaining 50 time points was generated using a stationary two-state Markov chain with a lag-1 persistence of 0.3, producing temporally clustered blocks of missing observations. Masks were redrawn when fewer than three follow-up observations remained. This design introduces non-random subject-level heterogeneity and consecutive missing periods, reflecting variations in patient adherence and clustered missed assessments commonly observed in longitudinal healthcare monitoring.

The class-level GP variance and length-scale were 1 and 100 in both scenarios. The individual-level values were 10 and 150 for the RBF scenario and 10 and 250 for the Matérn-5/2 scenario. Gaussian noise with $\sigma = 0.05$ was added in both settings, and GP-based methods used the corresponding kernel. Missing ratios ranged from 50% to 90%. For each ratio, we conducted 50 repetitions using 75 samples per class and a stratified 70/30 training and test split. Performance was evaluated using test AUC and MSE calculated only at masked time points.

Figure 1 illustrates the effects of increasing missing ratios under even and baseline-dependent Markov missingness. The left column shows the complete trajectories of two representative subjects from each class, with class 0 shown in blue and class 1 in red. The middle and right columns display the retained observations at missing ratios of 70% and 90%, respectively. As the missing ratio increases, fewer observations remain to support reconstruction of the underlying trajectories. Even missingness produces observations distributed across the study period, whereas Markov missingness generates temporally clustered patterns of observed and missing values.

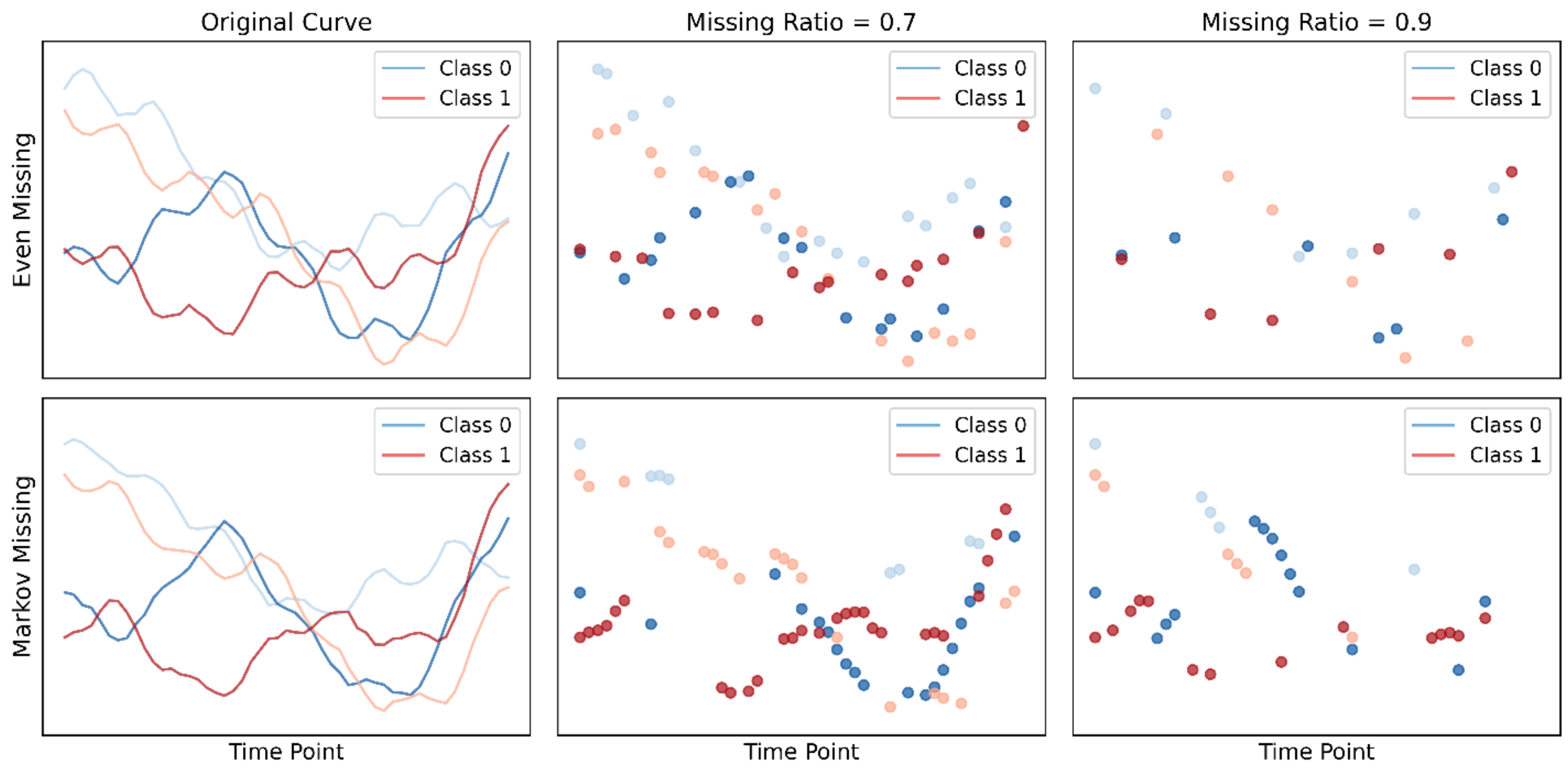

Figure 1. Visualization of simulated time series under even and baseline-dependent Markov missingness. The top and bottom rows represent even and Markov missingness, respectively. The left column shows the complete trajectories for two representative subjects from each class, with class 0 shown in blue and class 1 in red. The middle and right columns display the retained observations from the same trajectories at missing ratios of 70% and 90%, respectively.

### *5.2. Model performance comparison*

Table 1 presents the results under RBF data generation with evenly distributed missingness. MAGIC maintained an AUC of approximately 0.99 through the 80% missing ratio and achieved the highest AUC from 60% to 90%, while performing comparably to GRU-D at 50%. The slight increase in AUC through 80% missingness may reflect greater reliance on the learned class-specific mean processes, which preserve class-discriminative patterns when individual observations become sparse. However, the differences were small relative to the variability across repetitions and should therefore be interpreted cautiously. At 90% missingness, too few observations remained to reliably anchor trajectory reconstruction and class prediction, resulting in an AUC decrease to 0.8090. MAGIC also achieved the lowest MSE across all missing ratios.

Table 1. Performance comparison under RBF data generation with evenly distributed missingness. Values represent the mean and standard deviation in parentheses of test AUC and masked-point MSE over 50 repetitions. Bold entries indicate the best result at each missing ratio. MAGIC achieved the highest AUC at missing ratios from 60% to 90% and the lowest MSE across all missing ratios.

| AUC | | | | | |
|---|---|---|---|---|---|
| Missing Ratio | 50% | 60% | 70% | 80% | 90% |
| GRU-D | **0.9935 (0.0434)** | 0.9799 (0.0634) | 0.9826 (0.0265) | 0.8330 (0.1123) | 0.6297 (0.1090) |
| ODE-RNN | 0.9636 (0.0859) | 0.8864 (0.1252) | 0.8053 (0.1020) | 0.6723 (0.1051) | 0.5719 (0.1003) |
| STraTS | 0.6362 (0.1111) | 0.6376 (0.1099) | 0.6392 (0.1084) | 0.6296 (0.1084) | 0.6290 (0.1056) |
| NRTSI | 0.9878 (0.0138) | 0.9559 (0.0353) | 0.9175 (0.0493) | 0.8191 (0.0711) | 0.6942 (0.1012) |
| NRTSI-c22 | 0.5806 (0.0875) | 0.6026 (0.1131) | 0.6018 (0.0849) | 0.5970 (0.0810) | 0.6016 (0.0963) |
| SGP | 0.9117 (0.0600) | 0.8972 (0.0675) | 0.8413 (0.0814) | 0.7906 (0.0965) | 0.7024 (0.1002) |
| MTGP | 0.9412 (0.0463) | 0.9308 (0.0537) | 0.9340 (0.0537) | 0.8472 (0.0801) | 0.7010 (0.0979) |
| MAGIC-NC | 0.9510 (0.0402) | 0.9517 (0.0435) | 0.9796 (0.0209) | 0.9854 (0.0459) | 0.7441 (0.1015) |
| MAGIC | 0.9902 (0.0227) | **0.9912 (0.0208)** | **0.9928 (0.0229)** | **0.9929 (0.0315)** | **0.8090 (0.0982)** |
| MSE | | | | | |
| NRTSI | 0.9454 (0.2181) | 1.1233 (0.2414) | 1.7475 (0.4686) | 2.5507 (0.5605) | 4.5209 (1.1339) |
| SGP | 0.8126 (0.0586) | 1.0448 (0.0744) | 1.4040 (0.1107) | 2.0631 (0.4017) | 3.7793 (1.2207) |
| MTGP | 0.2065 (0.0943) | 0.2537 (0.1301) | 0.3335 (0.1959) | 0.6777 (0.4045) | 3.9941 (0.4575) |
| MAGIC-NC | 0.1633 (0.0645) | 0.1598 (0.0400) | 0.1986 (0.1297) | 0.6979 (0.3560) | 0.9523 (0.5871) |
| MAGIC | **0.0258 (0.0973)** | **0.0287 (0.0835)** | **0.0594 (0.1566)** | **0.2433 (0.2488)** | **0.7233 (0.5711)** |

Table 2 reports the results under Matérn-5/2 data generation with baseline-dependent Markov missingness. In this more complex setting, AUC declined consistently as missingness increased. Nevertheless, MAGIC achieved the highest AUC and lowest MSE at every missing ratio, retaining an AUC of 0.8554 at 90% missingness. Across both settings, MAGIC consistently outperformed MAGIC-NC in both AUC and MSE. This result supports the contribution of class-specific information to trajectory reconstruction and classification, particularly under severe and temporally clustered missingness.

Table 2. Performance comparison under Matérn-5/2 data generation with baseline-dependent Markov missingness. Values represent the mean and standard deviation in parentheses of test AUC and masked-point MSE over 50 repetitions. Bold entries indicate the best result at each missing ratio. MAGIC achieved the highest AUC and the lowest MSE across all missing ratios.

| AUC | | | | | |
|---|---|---|---|---|---|
| Missing Ratio | 50% | 60% | 70% | 80% | 90% |
| GRU-D | 0.7736 (0.1237) | 0.7000 (0.1168) | 0.6512 (0.0935) | 0.6004 (0.1021) | 0.5753 (0.1070) |
| ODE-RNN | 0.6488 (0.1237) | 0.6466 (0.0998) | 0.5995 (0.1019) | 0.5747 (0.1018) | 0.5777 (0.0924) |
| STraTS | 0.6185 (0.1175) | 0.6239 (0.1047) | 0.6106 (0.1106) | 0.6016 (0.1134) | 0.5972 (0.1045) |
| NRTSI | 0.8783 (0.0591) | 0.8438 (0.0702) | 0.7699 (0.0877) | 0.7227 (0.0905) | 0.6391 (0.0919) |
| NRTSI-c22 | 0.5834 (0.0875) | 0.5816 (0.1097) | 0.5796 (0.0920) | 0.5646 (0.1057) | 0.5578 (0.0959) |
| SGP | 0.9263 (0.0518) | 0.9025 (0.0491) | 0.8610 (0.0708) | 0.7944 (0.0840) | 0.7201 (0.0916) |
| MTGP | 0.9704 (0.0271) | 0.9634 (0.0269) | 0.9199 (0.0435) | 0.8603 (0.0585) | 0.7560 (0.0919) |
| MAGIC-NC | 0.9731 (0.0257) | 0.9657 (0.0260) | 0.9309 (0.0351) | 0.8777 (0.0567) | 0.7749 (0.0940) |
| MAGIC | **0.9919 (0.0132)** | **0.9897 (0.0152)** | **0.9739 (0.0303)** | **0.9469 (0.0331)** | **0.8554 (0.0600)** |
| MSE | | | | | |
| NRTSI | 2.1288 (1.1315) | 2.6076 (1.0696) | 3.5921 (1.2404) | 5.3252 (1.5423) | 8.9077 (4.3612) |
| SGP | 0.7999 (0.2655) | 1.0751 (0.3991) | 1.5455 (0.4008) | 2.1414 (0.4589) | 3.1562 (0.5548) |
| MTGP | 0.4622 (0.2561) | 0.7031 (0.4076) | 1.1880 (0.4393) | 1.6509 (0.4488) | 2.5343 (0.5638) |
| MAGIC-NC | 0.4034 (0.2285) | 0.6174 (0.3656) | 1.0090 (0.3904) | 1.4038 (0.3678) | 2.1530 (0.3731) |
| MAGIC | **0.1659 (0.1366)** | **0.3079 (0.2027)** | **0.5394 (0.2602)** | **0.9968 (0.3166)** | **2.0032 (0.4457)** |

The imputation results followed a similar pattern. As shown in Figure 2, at 80% missingness, MAGIC more closely followed the original trajectories than the competing methods, preserving their overall shapes and major temporal variations while avoiding irregular fluctuations. The right panels further demonstrate that MAGIC closely recovered the original trajectories at 60% and 80% missingness and retained their primary temporal patterns even at 90% missingness, although deviations increased under this extreme sparsity. These examples visually support the MSE results and demonstrate MAGIC's ability to reconstruct underlying trajectories under severe and temporally clustered missingness.

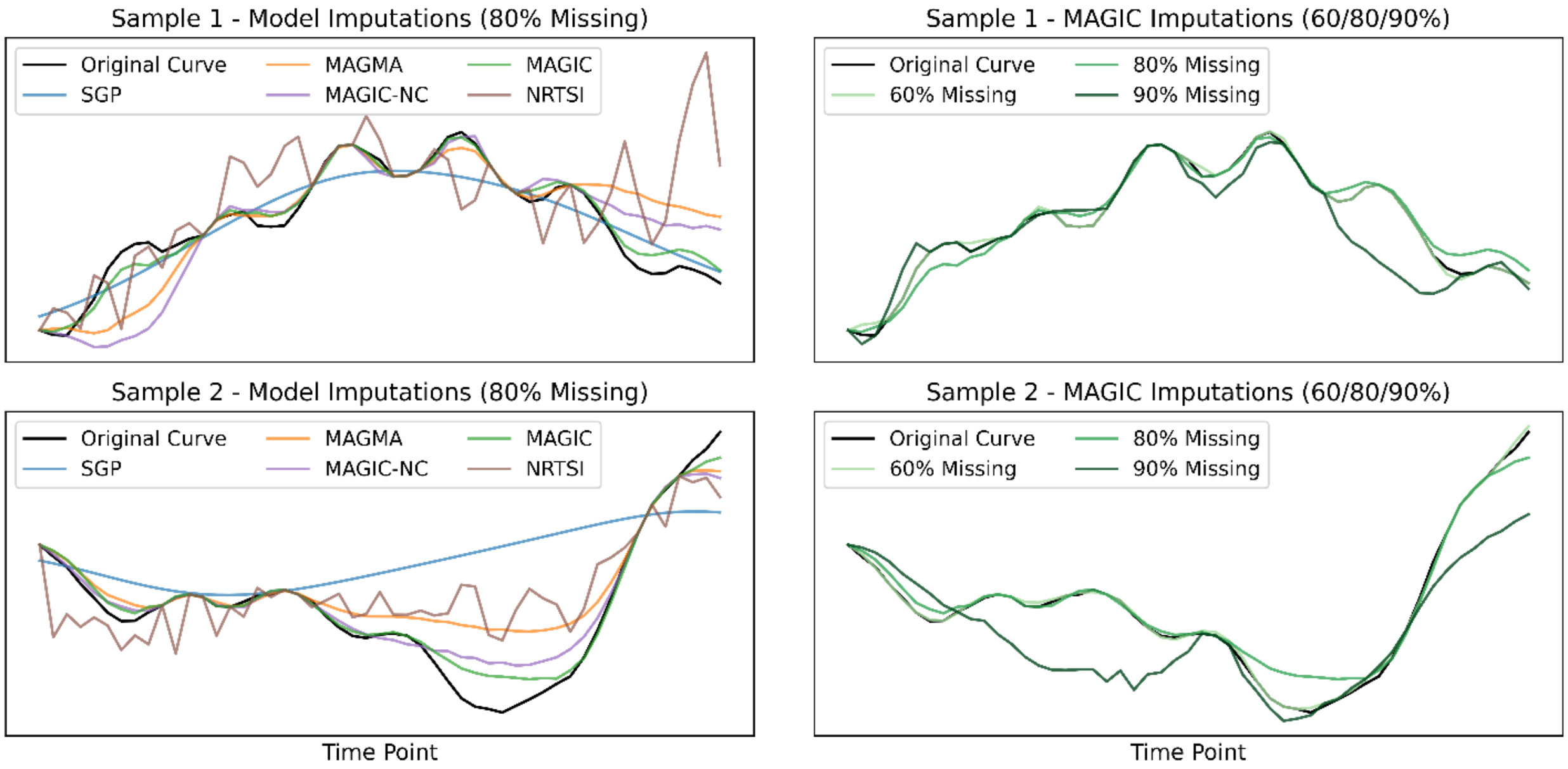

Figure 2. Visualization of imputation performance for two representative time series, with Sample 1 in the top row and Sample 2 in the bottom row. The left panels compare the original complete trajectories with those imputed by SGP, MAGMA, MAGIC-NC, MAGIC, and NRTSI at an 80% missing ratio. The right panels compare the original trajectories with MAGIC imputations at missing ratios of 60%, 80%, and 90%.

### 5.3. Sensitivity, robustness, and computational analyses

#### 5.3.1. Smoothing-penalty sensitivity

We evaluated the sensitivity of MAGIC to the smoothing-scale parameter $\rho$ under the Matérn-5/2 data-generating process with baseline-dependent Markov missingness. We considered $\rho \in \{0.02, 0.05, 0.1, 0.2, 0.5, 1\}$, together with a no-penalty setting, across missing ratios from 50% to 90%. As the effective roughness-penalty weight is $\rho^{-4}$, smaller values of $\rho$ impose stronger smoothing, whereas larger values correspond to weaker regularization. Figure 3 reports the mean AUC and MSE across 50 simulation repetitions. Overall, $\rho = 0.2$ provided the best balance between classification and trajectory reconstruction, producing the highest AUC and lowest or nearly lowest MSE across the evaluated missing ratios. The MAGIC results reported in Table 2 were obtained using ρ=0.2.

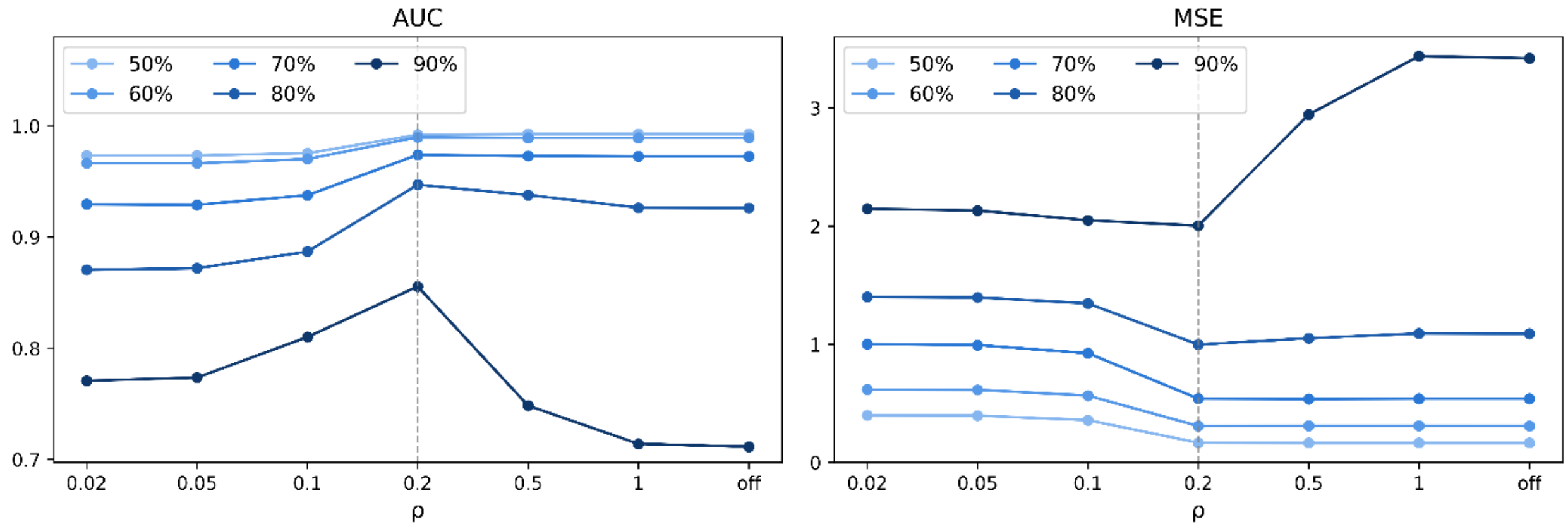

Figure 3. Sensitivity of classification and imputation performance to the smoothing-scale parameter under missing ratios ranging from 50% to 90%. The left and right panels report the mean AUC and MSE, respectively, across 50 simulation repetitions. Smaller values of correspond to stronger smoothing. The dashed vertical line identifies the smoothing level that maximized AUC and minimized or nearly minimized MSE across the evaluated missing ratios.

Figure 4 further illustrates how the smoothing level affects estimation of the class-specific mean functions. Without the penalty, or when the penalty was weak, the fitted mean curves exhibited substantial local fluctuations caused by overfitting to sparse observations. In contrast, excessively strong smoothing, such as that produced by $\rho = 0.05$, suppressed meaningful temporal variation and resulted in fitted curves that were nearly linear. Both insufficient and excessive smoothing consequently degraded AUC and increased MSE, as shown in Figure 3. At $\rho = 0.2$, the fitted curves retained the major temporal patterns of the true class-level mean functions while avoiding irregular fluctuations, explaining its superior classification and imputation performance.

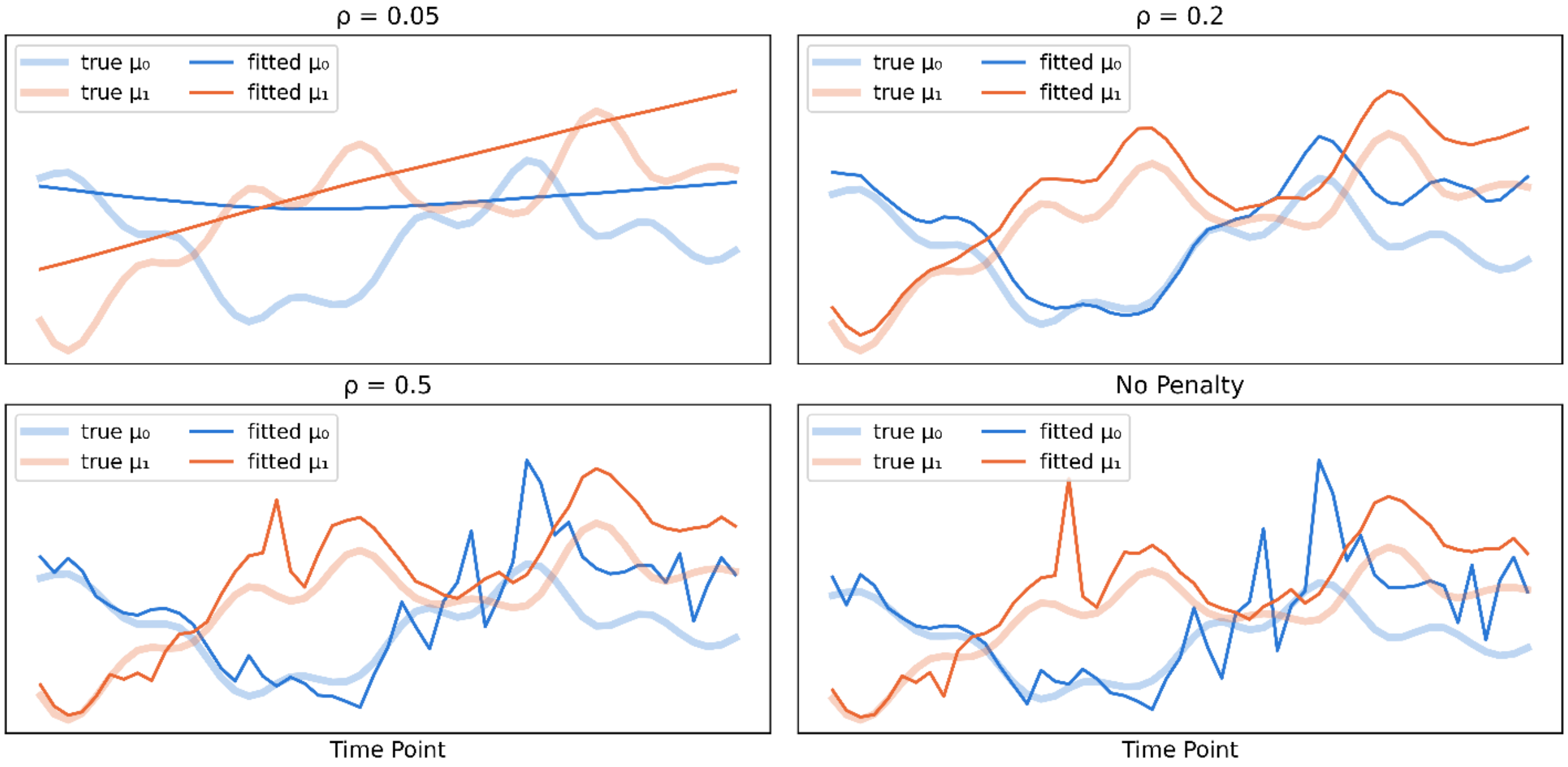

Figure 4. Estimated class-level mean functions under different levels of smoothing. The light blue and light red curves represent the true class-specific mean functions for classes 0 and 1, respectively, while the darker curves represent their fitted counterparts. The panels show the fitted mean functions at $\rho$ of 0.05, 0.2, and 0.5, together with the no-penalty setting. Smaller parameter values impose stronger smoothing, whereas larger values produce weaker regularization.

#### 5.3.2. Robustness to model misspecification

The RBF kernel was initially selected because it provides a smooth stationary prior, whereas the Matérn-5/2 kernel permits less restrictive smoothness. To assess sensitivity to this choice, we fitted MAGIC using Matérn-5/2 kernels for the RBF-generated data and RBF kernels for the Matérn-5/2-generated data. We also replaced the Gaussian sampling distribution in the Matérn-5/2 Markov setting with Student's $t$ distributions with three and five degrees of freedom while retaining the Gaussian Matérn-5/2 model for fitting. These experiments considered missing ratios of 70%, 80%, and 90%, with all other settings unchanged.

Table 3 shows that the Matérn-5/2 fit improved both AUC and MSE for the RBF-generated data. For the Matérn-5/2-generated data, the alternative RBF fit retained high AUC through 80% missingness but showed greater degradation at 90%, with an AUC of 0.7709 compared with 0.8554 under the matched kernel. Under the non-Gaussian settings, AUC remained comparable to the Gaussian result across all missing ratios. MSE became larger and more variable at 80% and 90%, particularly under the heavier-tailed $t_3$ distribution. These findings indicate that MAGIC's classification performance is not dependent on the exact RBF specification, although an appropriate kernel remains important for accurate reconstruction under severe missingness.

Table 3. Robustness of MAGIC under alternative fitted kernels and non-Gaussian data generation. Values represent the mean and standard deviation in parentheses of test AUC and masked-point MSE over 50 repetitions. $t_\nu$ denotes a Student's distribution with $\nu$ degrees of freedom.

| | | | AUC | | | MSE | | |
|---|---|---|---|---|---|---|---|---|
| Data Generation | Sampling Distribution | Fitted kernel | 70% | 80% | 90% | 70% | 80% | 90% |
| RBF, even | Gaussian | RBF | 0.9928 (0.0229) | 0.9929 (0.0315) | 0.8090 (0.0982) | 0.0594 (0.1566) | 0.2433 (0.2488) | 0.7233 (0.5711) |
| | Gaussian | Matérn-5/2 | 0.9998 (0.0011) | 0.9985 (0.0089) | 0.8594 (0.0867) | 0.0307 (0.0379) | 0.0927 (0.1087) | 0.4829 (0.2681) |
| Matérn-5/2, Markov | Gaussian | Matérn-5/2 | 0.9739 (0.0303) | 0.9469 (0.0331) | 0.8554 (0.0600) | 0.5394 (0.2602) | 0.9968 (0.3166) | 2.0032 (0.4457) |
| | Gaussian | RBF | 0.9700 (0.0304) | 0.9232 (0.0480) | 0.7709 (0.0842) | 0.9369 (0.3933) | 1.8130 (0.4957) | 5.0291 (2.1984) |
| | $t_5$ | Matérn-5/2 | 0.9749 (0.0227) | 0.9526 (0.0376) | 0.8558 (0.0756) | 0.5035 (0.2200) | 1.1393 (0.4576) | 2.6091 (0.8268) |
| | $t_3$ | Matérn-5/2 | 0.9775 (0.0245) | 0.9504 (0.0347) | 0.8700 (0.0675) | 0.4851 (0.2085) | 1.3500 (1.7029) | 2.8887 (2.2564) |

### 5.3.3. Computational scalability and optimization stability

We evaluated the computational scalability and convergence behavior of MAGIC under 70% missingness by varying the number of time points and sample size. Table 4 summarizes the results across 50 independent repetitions for each configuration. No convergence failures were observed, and the mean number of EM iterations ranged from 19.1 to 23.8. Increasing the number of time points from 50 to 100 increased the mean fitting time from 16.6–19.1 seconds to 55.3–60.6 seconds, reflecting the computational cost of the GP covariance operations. In comparison, increasing the sample size produced a relatively modest increase in runtime. Peak memory usage remained stable across the evaluated configurations, ranging from approximately 502 to 512 MB.

We also examined sensitivity to initialization using a fixed dataset generated with data seed 0 under the Matérn-5/2 setting with baseline-dependent Markov missingness at a 70% missing ratio. Only the initial hyperparameter values were varied across 50 initialization seeds. AUC remained unchanged to four decimal places at 0.9842, while masked MSE had a mean of 0.1923 and a standard deviation of 0.0002, ranging from 0.1916 to 0.1927. These results indicate that predictive and imputation performance was stable across the tested hyperparameter initializations. Numerical stability was further supported by bounded L-BFGS-B optimization and the monotonicity safeguard, which accepts a block update only when it increases the Q-function.

Table 4. Computational scalability and convergence behavior of MAGIC across different numbers of time points and sample sizes under 70% missingness. Results are based on 50 independent simulation repetitions. "Fail" denotes the number of runs that did not satisfy the convergence criterion, and "Min–Max" gives the observed range of EM iterations. Iteration count, fitting time, and peak resident set size (RSS) are reported as the mean, with the standard deviation in parentheses.

| Time Points | Sample Size | Fail | Iteration | Min–Max | Fit Time (s) | Peak RSS (MB) |
|---|---|---|---|---|---|---|
| 50 | 50 | 0 | 23.8 (5.6) | 17–51 | 16.6 (2.7) | 505.2 (3.4) |
| 100 | 50 | 0 | 19.1 (4.3) | 13–29 | 55.3 (12.9) | 510.4 (3.7) |
| 50 | 75 | 0 | 22.9 (3.8) | 16–33 | 17.4 (2.5) | 509.1 (2.7) |
| 100 | 75 | 0 | 20.6 (5.1) | 13–34 | 58.8 (13.6) | 502.2 (0.3) |
| 50 | 100 | 0 | 23.2 (4.6) | 17–40 | 19.1 (3.1) | 511.5 (2.3) |
| 100 | 100 | 0 | 21.0 (5.0) | 14–34 | 60.6 (12.9) | 508.7 (4.4) |

# 6. Case study

## *6.1. PTH recovery prediction*

### *6.1.1. Data collection and modelling*

The study cohort included 50 participants with mild traumatic brain injury (mTBI) and post-traumatic headache (PTH) enrolled in a prospective study with Institutional Review Board (IRB) approval. As part of this study, participants completed an electronic speech application on their mobile devices prompting them to read aloud five standardized sentences every three days over a period of three months. The robust Voice Activity Detection (rVAD) algorithm was utilized to identify the start and end of speech (Tan et al., 2020). Speaking rate was computed as the number of syllables divided by speaking time and normalized using a sex- and age-matched control cohort from the Mozilla database (Ardila et al., 2020). Headache intensity was recorded on a 0-9 scale (0 indicating no headache and 9 indicating worst possible headache) prior to beginning the speech task. In addition, PTH improvement status was assessed for each participant at three months post-mTBI according to established criteria (Mao et al., 2023; Ku et al., 2025). 28 participants were classified as having PTH improvement and 22 as having no PTH improvement.

Despite this structured design, participants often missed their speech tasks, resulting in an average missing ratio of 71.9%. Given this data sparsity and limited sample size, we hypothesized that MAGIC would provide a suitable model to predict three-month PTH improvement status using the time series of speaking rate and headache intensity. We utilized only the first six weeks of data post-mTBI to predict PTH improvement status. This approach was designed to test the feasibility of early prediction, which could enable more timely clinical interventions.

We applied MAGIC separately to the speaking rate and headache intensity time series. For comparison, we evaluated GRU-D, ODE-RNN, STraTS, NRTSI, NRTSI-C22, SGP, MTGP, and MAGIC-NC under the same validation scheme. Combined predictions were constructed according to each model's design. GRU-D, ODE-RNN, and STraTS jointly used both time series as multivariate inputs for classification. For MAGIC, MAGIC-NC, SGP, MTGP, NRTSI, and NRTSI-C22, the two time series were modeled separately, and their predicted probabilities were averaged for each test participant before calculating AUC. Leave-one-out cross-validation (LOOCV) was used to assess prediction accuracy. To evaluate imputation accuracy for methods producing reconstructed trajectories, we employed a nested cross-validation structure. For each test sample in the outer LOOCV, we applied an inner LOOCV that masked one existing value, imputed it, and calculated MSE against the original value. The mean and standard deviation of MSE were then calculated across all samples.

### *6.1.2. Model performance comparison*

Table 5 summarizes the model performance. MAGIC achieved the highest AUC using speaking rate, headache intensity, and their combination, with AUC values of 0.7013, 0.7451, and 0.8052, respectively. MAGIC also achieved the lowest speaking-rate MSE, while NRTSI achieved the lowest headache-intensity MSE. MAGIC outperformed MAGIC-NC across all reported AUC and MSE measures, further supporting the contribution of class-specific information.

Table 5. Model comparison for PTH improvement prediction using speaking rate and headache intensity time series. AUC was evaluated using leave-one-out cross-validation, and MSE is reported as mean with standard deviation in parentheses across samples using nested leave-one-out cross-validation. Combined denotes the average of the feature-specific predicted probabilities.

| Metrics | AUC | | | MSE | |
|---|---|---|---|---|---|
| Models | Speaking Rate | Headache Intensity | Combined | Speaking Rate | Headache Intensity |
| GRU-D | 0.6558 | 0.7029 | 0.7857 | - | - |
| ODE-RNN | 0.6494 | 0.6769 | 0.7532 | - | - |
| STraTS | 0.6753 | 0.7419 | 0.6786 | - | - |
| NRTSI | 0.6867 | 0.7354 | 0.7549 | 0.0226 (0.0151) | **3.0434 (3.4459)** |
| NRTSI-c22 | 0.6494 | 0.5130 | 0.6169 | | |
| SGP | 0.6948 | 0.7045 | 0.7646 | 0.0291 (0.0345) | 4.2105 (5.9327) |
| MTGP | 0.6867 | 0.7094 | 0.7370 | 0.0225 (0.0206) | 4.4360 (7.8916) |
| MAGIC-NC | 0.6981 | 0.7175 | 0.7597 | 0.0336 (0.0260) | 4.0563 (5.2980) |
| MAGIC | **0.7013** | **0.7451** | **0.8052** | **0.0202 (0.0243)** | 3.2787 (4.2526) |

## ***6.2.ICU mortality prediction***

### *6.2.1. Data collection and modelling*

The analysis utilized the publicly available PhysioNet Challenge 2012 dataset, which comprises records from 12,000 ICU admissions. Our primary objective was to predict in-hospital mortality as the response variable. Initial data preprocessing involved replacing negative values with null entries and analyzing temporal measurement patterns, which revealed that most measurements were recorded at 30 minutes or hourly intervals. For feature screening, we first aggregated each of the 37 variables into one-hour intervals and used all eligible admissions with in-hospital death and an equal number of randomly selected survivors. Each variable was evaluated separately using MAGIC, MAGMA, and SGP, and the three AUC values were averaged to obtain a common selection score. The five highest-ranking variables were BUN [Blood urea nitrogen], Creatinine [Serum creatinine], GCS [Glasgow Coma Score], Urine [Urine output], and RespRate [Respiration rate]. After feature selection, we reconstructed the five selected variables using 30-minute bins across the 48-hour observation period, yielding 97 discrete time points. When more than one measurement occurred within a 30-minute interval, the values were averaged. The final cohort included 2,927 patient admissions with at least one measurement recorded across five selected features, consisting of 2,698 survivors and 229 in-hospital deaths. The corresponding missing percentages were 96.69% (BUN), 96.68% (Creatinine), 83.79% (GCS), 70.85% (Urine), and 53.15% (RespRate).

To assess the model performance under a limited sample size, we randomly chose 100 admissions from each class (survivor vs. in-hospital death) and split them into 70% training and 30% test sets. The same competing methods described in Section 6.1.1 were evaluated. Min-max standardization was applied to the training set, and the same scaling parameters were used for the test set. To evaluate performance when individual admissions were highly sparse, we increased the missing ratio to 80% separately for each admission for GCS, Urine, and RespRate. For admissions with missingness below 80%, observed values were randomly masked until the target ratio was reached. For all five features, imputation accuracy was instead evaluated using LOOCV within each test admission, where each available observation was withheld in turn and compared with its imputed value. Combined predictions were constructed according to each model's design. Following the same model-specific procedure as in Section 6.1.1, GRU-D, ODE-RNN, and STraTS jointly used all five time series as multivariate inputs. For the remaining feature-wise methods, the five predicted probabilities were averaged for each test admission before calculating AUC. This procedure was repeated 50 times, and the mean and standard deviation of test AUC and MSE were reported.

### *6.2.2. Model performance comparison*

Table 6 summarizes the model performance. MAGIC achieved the highest mean AUC for four of the five individual features and for the combined analysis, while ODE-RNN achieved the highest AUC for GCS. Averaging the feature-specific probabilities produced a combined AUC of 0.8332 for MAGIC. MAGIC also achieved the lowest imputation MSE for all five features. In particular, MAGIC achieved both the highest AUC and lowest MSE for BUN and Creatinine, which had the highest missing ratios.

Table 6. Model comparison for ICU mortality prediction using five clinical time series. Values represent the mean and standard deviation in parentheses of test AUC and imputation MSE over 50 repetitions.

| AUC | | | | | | |
|---|---|---|---|---|---|---|
| Models | BUN | Creatinine | GCS | Urine | RespRate | Combined |
| GRU-D | 0.6939 (0.1179) | 0.6263 (0.1084) | 0.6971 (0.0735) | 0.7518 (0.0690) | 0.5506 (0.0813) | 0.8062 (0.0601) |
| ODE-RNN | 0.7161 (0.0737) | 0.6107 (0.1026) | **0.7021 (0.0717)** | 0.7524 (0.0748) | 0.5487 (0.0799) | 0.8151 (0.0529) |
| STraTS | 0.7519 (0.0602) | 0.6787 (0.0672) | 0.6909 (0.0785) | 0.7510 (0.0695) | 0.6091 (0.0726) | 0.8176 (0.0551) |
| NRTSI | 0.7614 (0.0632) | 0.6765 (0.0750) | 0.6963 (0.0731) | 0.7570 (0.0674) | 0.6290 (0.0688) | 0.8276 (0.0543) |
| NRTSI-c22 | 0.6783 (0.0786) | 0.6173 (0.0762) | 0.6215 (0.0893) | 0.6458 (0.0735) | 0.5448 (0.0682) | 0.7400 (0.0689) |
| SGP | 0.7328 (0.0665) | 0.6516 (0.0882) | 0.6644 (0.0747) | 0.7589 (0.0667) | 0.5953 (0.0721) | 0.8054 (0.0585) |
| MTGP | 0.7304 (0.0696) | 0.6446 (0.0980) | 0.6924 (0.0649) | 0.7645 (0.0638) | 0.6212 (0.0633) | 0.8163 (0.0565) |
| MAGIC-NC | 0.7660 (0.0624) | 0.6750 (0.0918) | 0.7005 (0.0715) | 0.7676 (0.0636) | 0.6251 (0.0628) | 0.8308 (0.0552) |
| MAGIC | **0.7694 (0.0635)** | **0.6795 (0.0916)** | 0.6954 (0.0708) | **0.7693 (0.0623)** | **0.6353 (0.0642)** | **0.8332 (0.0562)** |
| MSE | | | | | | |
| NRTSI | 0.0119 (0.0073) | 0.0305 (0.1248) | 0.0113 (0.0050) | 0.0262 (0.1433) | 0.0066 (0.0023) | - |
| SGP | 0.0305 (0.0210) | 0.0187 (0.0147) | 0.0938 (0.1543) | 0.0052 (0.0032) | 0.0078 (0.0021) | - |
| MTGP | 0.0207 (0.0124) | 0.0187 (0.0169) | 0.0100 (0.0065) | 0.0046 (0.0028) | 0.0063 (0.0022) | - |
| MAGIC-NC | 0.0076 (0.0031) | 0.0062 (0.0050) | 0.0084 (0.0030) | 0.0044 (0.0028) | 0.0060 (0.0021) | - |
| MAGIC | **0.0047 (0.0025)** | **0.0050 (0.0044)** | **0.0075 (0.0031)** | **0.0042 (0.0027)** | **0.0057 (0.0021)** | - |

To assess whether the findings were sensitive to the original sample size, we conducted an expanded-sample analysis using all 229 eligible admissions with in-hospital death and an equal number of randomly selected survivors. The same 70/30 training and test split was repeated 50 times, and this analysis focused on combined AUC as the classification outcome integrating all five features. As shown in Table 7, MAGIC retained the highest mean combined AUC of 0.8302, followed by MAGIC-NC at 0.8282 and NRTSI at 0.8234. The result was consistent with the combined AUC of 0.8332 obtained using 100 admissions per class, indicating that the primary performance conclusion was not driven by the original subsample size.

Table 7. Combined AUC in the expanded-sample ICU mortality analysis using 229 admissions per class, comprising all eligible in-hospital deaths and an equal number of randomly selected survivors. Values represent the mean and standard deviation in parentheses over 50 repetitions.

| Model | AUC | Model | AUC |
|---|---|---|---|
| GRU-D | 0.7983 (0.0388) | SGP | 0.8069 (0.0399) |
| ODE-RNN | 0.8120 (0.0356) | MTGP | 0.8220 (0.0357) |
| STraTS | 0.8216 (0.0390) | MAGIC-NC | 0.8282 (0.0355) |
| NRTSI | 0.8234 (0.0348) | MAGIC | **0.8302 (0.0354)** |

## 7. Conclusion

This paper introduces MAGIC, a novel supervised hierarchical MTGP framework for joint missing value imputation and time series classification. Rather than treating imputation and classification as independent stages, MAGIC leverages class-specific information in a unified framework, optimizing both imputation and classification tasks concurrently through an integrated parameter estimation process. Experimental results demonstrated MAGIC's superior performance, particularly in scenarios with high missing ratios. The model's performance was further validated through two healthcare applications: PTH recovery prediction and ICU mortality prediction, where it achieved the best overall performance across most evaluation metrics.

There are several limitations of this study, which point out some directions for future research. First, the current formulation treats features independently and aggregates feature-specific predicted probabilities rather than directly modeling inter-feature dependencies. Extending MAGIC to a multivariate framework could capture cross-feature interactions and further improve both imputation accuracy and classification performance. Second, the framework is constrained to binary classification. Generalizing MAGIC to handle multiclass classification problems or continuous regression tasks would broaden its applicability to more complex healthcare applications. Third, although the estimation procedure accounts for uncertainty in the latent class-level mean functions, the functional logistic regression uses posterior-mean trajectories as plug-in inputs and may therefore produce overly confident predictions under severe missingness.

## Acknowledgements

Funding sources are hidden due to double-blind requirements.

## Data availability statement

Data availability statement is hidden due to double-blind requirements.

## References

Aguilera-Morillo, M. C., Aguilera, A. M., Escabias, M., & Valderrama, M. J. (2013). Penalized spline approaches for functional logit regression. *TEST*, *22*(2), 251–277. https://doi.org/10.1007/s11749-012-0307-1

Alaa, A. M., & Van Der Schaar, M. (2017). Bayesian inference of individualized treatment effects using multi-task gaussian processes. *Advances in Neural Information Processing Systems*, *30*.

Ardila, R., Branson, M., Davis, K., Henretty, M., Kohler, M., Meyer, J., Morais, R., Saunders, L., Tyers, F. M., & Weber, G. (2020). *Common Voice: A Massively-Multilingual Speech Corpus* (arXiv:1912.06670). arXiv. https://doi.org/10.48550/arXiv.1912.06670

Ashina, H., Al-Khazali, H. M., Iljazi, A., Ashina, S., Amin, F. M., Lipton, R. B., & Schytz, H. W. (2021). Psychiatric and cognitive comorbidities of persistent post-traumatic headache attributed to mild traumatic brain injury. *The Journal of Headache and Pain*, *22*(1), 83. https://doi.org/10.1186/s10194-021-01287-7

Aydin, S. (2022). Time Series Analysis and Some Applications in Medical Research. *Journal of Mathematics and Statistics Studies*, *3*(2), 31–36. https://doi.org/10.32996/jmss.2022.3.2.3

Banerjee, S., & Gelfand, A. E. (2002). Prediction, interpolation and regression for spatially misaligned data. *Sankhyā: The Indian Journal of Statistics, Series A*, 227–245.

Bashir, F., & Wei, H.-L. (2018). Handling missing data in multivariate time series using a vector autoregressive model-imputation (VAR-IM) algorithm. *Neurocomputing*, *276*, 23–30. https://doi.org/10.1016/j.neucom.2017.03.097

Bonilla, E. V., Chai, K., & Williams, C. (2007). Multi-task Gaussian process prediction. *Advances in Neural Information Processing Systems*, *20*.

Chaudhry, S. I., Mattera, J. A., Curtis, J. P., Spertus, J. A., Herrin, J., Lin, Z., Phillips, C. O., Hodshon, B. V., Cooper, L. S., & Krumholz, H. M. (2010). Telemonitoring in Patients with Heart Failure. *New England Journal of Medicine*, *363*(24), 2301–2309. https://doi.org/10.1056/NEJMoa1010029

Che, Z., Purushotham, S., Cho, K., Sontag, D., & Liu, Y. (2018). Recurrent neural networks for multivariate time series with missing values. *Scientific Reports*, *8*(1), 6085.

Chong, C. D., Zhang, J., Li, J., Wu, T., Dumkrieger, G., Nikolova, S., Ross, K., Stegmann, G., Liss, J., Schwedt, T. J., Jayasuriya, S., & Berisha, V. (2021). Altered speech patterns in subjects with post-traumatic headache due to mild traumatic brain injury. *The*

*Journal of Headache and Pain*, *22*(1), 82. https://doi.org/10.1186/s10194-021-01296-6

Denhere, M., & Billor, N. (2016). Robust Principal Component Functional Logistic Regression. *Communications in Statistics - Simulation and Computation*, *45*(1), 264–281. https://doi.org/10.1080/03610918.2013.861628

Escabias, M., Aguilera, A. M., & Valderrama, M. J. (2004). Principal component estimation of functional logistic regression: Discussion of two different approaches. *Journal of Nonparametric Statistics*, *16*(3–4), 365–384. https://doi.org/10.1080/10485250310001624738

Escabias, M., Aguilera, A. M., & Valderrama, M. J. (2007). Functional PLS logit regression model. *Computational Statistics & Data Analysis*, *51*(10), 4891–4902. https://doi.org/10.1016/j.csda.2006.08.011

Futoma, J., Hariharan, S., & Heller, K. (2017). Learning to detect sepsis with a multitask Gaussian process RNN classifier. *International Conference on Machine Learning*, 1174–1182.

Hanley, J., Ure, J., Pagliari, C., Sheikh, A., & McKinstry, B. (2013). Experiences of patients and professionals participating in the HITS home blood pressure telemonitoring trial: A qualitative study. *BMJ Open*, *3*(5), e002671. https://doi.org/10.1136/bmjopen-2013-002671

Harezlak, J., Coull, B. A., Laird, N. M., Magari, S. R., & Christiani, D. C. (2007). Penalized solutions to functional regression problems. *Computational Statistics & Data Analysis*, *51*(10), 4911–4925. https://doi.org/10.1016/j.csda.2006.09.034

Hayashi, K., Takenouchi, T., Tomioka, R., & Kashima, H. (2012). Self-measuring Similarity for Multi-task Gaussian Process. *Transactions of the Japanese Society for Artificial Intelligence*, *27*(3), 103–110. https://doi.org/10.1527/tjsai.27.103

Hori, T., Montcho, D., Agbangla, C., Ebana, K., Futakuchi, K., & Iwata, H. (2016). Multi-task Gaussian process for imputing missing data in multi-trait and multi-environment trials. *Theoretical and Applied Genetics*, *129*(11), 2101–2115. https://doi.org/10.1007/s00122-016-2760-9

Kar, S., Chawla, R., Haranath, S. P., Ramasubban, S., Ramakrishnan, N., Vaishya, R., Sibal, A., & Reddy, S. (2021). Multivariable mortality risk prediction using machine learning for COVID-19 patients at admission (AICOVID). *Scientific Reports*, *11*(1), 12801. https://doi.org/10.1038/s41598-021-92146-7

Kauppi, A. M., Edin, A., Ziegler, I., Mölling, P., Sjöstedt, A., Gylfe, Å., Strålin, K., & Johansson, A. (2016). Metabolites in Blood for Prediction of Bacteremic Sepsis in the Emergency Room. *PLOS ONE*, *11*(1), e0147670. https://doi.org/10.1371/journal.pone.0147670

Kaushik, S., Choudhury, A., Sheron, P. K., Dasgupta, N., Natarajan, S., Pickett, L. A., & Dutt, V. (2020). AI in Healthcare: Time-Series Forecasting Using Statistical, Neural, and Ensemble Architectures. *Frontiers in Big Data*, *3*, 4. https://doi.org/10.3389/fdata.2020.00004

Kazijevs, M., & Samad, M. D. (2023). Deep imputation of missing values in time series health data: A review with benchmarking. *Journal of Biomedical Informatics*, *144*, 104440. https://doi.org/10.1016/j.jbi.2023.104440

Kim, H., & Kim, H. (2018). Functional logistic regression with fused lasso penalty. *Journal of Statistical Computation and Simulation*, *88*(15), 2982–2999. https://doi.org/10.1080/00949655.2018.1491975

Kim, S., Kim, H., Yun, E., Lee, H., Lee, J., & Lee, J. (2023). Probabilistic imputation for time-series classification with missing data. *International Conference on Machine Learning*, 16654–16667.

Kishore, K., Braitberg, G., Holmes, N. E., & Bellomo, R. (2023). Early prediction of hospital admission of emergency department patients. *Emergency Medicine Australasia*, *35*(4), 572–588. https://doi.org/10.1111/1742-6723.14169

Ku, D., Mao, L., Nikolova, S., Dumkrieger, G. M., Ross, K. B., Huentelman, M., Anderson, T., Porreca, F., Navratilova, E., Starling, A., Wu, T., Li, J., Chong, C. D., & Schwedt, T. J. (2025). Longitudinal analysis of pain-induced brain activations in post-traumatic headache. *Cephalalgia*, *45*(5), 03331024251345160. https://doi.org/10.1177/03331024251345160

Leroy, A., Latouche, P., Guedj, B., & Gey, S. (2022). MAGMA: Inference and prediction using multi-task Gaussian processes with common mean. *Machine Learning*, *111*(5), 1821–1849. https://doi.org/10.1007/s10994-022-06172-1

Leroy, A., Latouche, P., Guedj, B., & Gey, S. (2023). Cluster-specific predictions with multi-task Gaussian processes. *Journal of Machine Learning Research*, *24*(5), 1–49.

Li, G., Shen, H., & Huang, J. Z. (2016). Supervised Sparse and Functional Principal Component Analysis. *Journal of Computational and Graphical Statistics*, *25*(3), 859–878. https://doi.org/10.1080/10618600.2015.1064434

Li, S. C.-X., & Marlin, B. M. (2016). A scalable end-to-end gaussian process adapter for irregularly sampled time series classification. *Advances in Neural Information Processing Systems*, *29*.

Lubba, C. H., Sethi, S. S., Knaute, P., Schultz, S. R., Fulcher, B. D., & Jones, N. S. (2019). Catch22: CAnonical Time-series CHaracteristics: CH Lubba et al. *Data Mining and Knowledge Discovery*, *33*(6), 1821–1852.

Mao, L., Dumkrieger, G., Ku, D., Ross, K., Berisha, V., Schwedt, T. J., Li, J., & Chong, C. D. (2023). Developing multivariable models for predicting headache improvement in patients with acute post-traumatic headache attributed to mild traumatic brain injury: A preliminary study. *Headache: The Journal of Head and Face Pain*, *63*(1), 136–145. https://doi.org/10.1111/head.14450

Meng, X.-L., & Rubin, D. B. (1993). Maximum likelihood estimation via the ECM algorithm: A general framework. *Biometrika*, *80*(2), 267–278.

Meystre, S. (2005). The Current State of Telemonitoring: A Comment on the Literature. *Telemedicine and E-Health*, *11*(1), 63–69. https://doi.org/10.1089/tmj.2005.11.63

Moritz, S., Sardá, A., Bartz-Beielstein, T., Zaefferer, M., & Stork, J. (2015). Comparison of different methods for univariate time series imputation in R. *arXiv Preprint arXiv:1510.03924*.

Morris, J. S. (2015). Functional Regression. *Annual Review of Statistics and Its Application*, *2*(1), 321–359. https://doi.org/10.1146/annurev-statistics-010814-020413

Nocedal, J. (1980). Updating quasi-Newton matrices with limited storage. *Mathematics of Computation*, *35*(151), 773–782.

Papini, S., Pisner, D., Shumake, J., Powers, M. B., Beevers, C. G., Rainey, E. E., Smits, J. A. J., & Warren, A. M. (2018). Ensemble machine learning prediction of posttraumatic stress disorder screening status after emergency room hospitalization. *Journal of Anxiety Disorders*, *60*, 35–42. https://doi.org/10.1016/j.janxdis.2018.10.004

Pare, G., Jaana, M., & Sicotte, C. (2007). Systematic Review of Home Telemonitoring for Chronic Diseases: The Evidence Base. *Journal of the American Medical Informatics Association*, *14*(3), 269–277. https://doi.org/10.1197/jamia.M2270

Piccialli, F., Giampaolo, F., Prezioso, E., Camacho, D., & Acampora, G. (2021). Artificial intelligence and healthcare: Forecasting of medical bookings through multi-source time-series fusion. *Information Fusion*, *74*, 1–16. https://doi.org/10.1016/j.inffus.2021.03.004

Rahman, M. A., Honan, B., Glanville, T., Hough, P., & Walker, K. (2020). Using data mining to predict emergency department length of stay greater than 4 hours: Derivation and single-site validation of a decision tree algorithm. *Emergency Medicine Australasia*, *32*(3), 416–421. https://doi.org/10.1111/1742-6723.13421

Ramsay, J. O., & Silverman, B. W. (2006). *Functional data analysis* (2. ed., [Nachdr.]). Springer.

Raso, M. G., Arcuri, F., Liperoti, S., Mercurio, L., Mauro, A., Cusato, F., Romania, L., Serra, S., Pignolo, L., Tonin, P., & Cerasa, A. (2021). Telemonitoring of Patients With Chronic Traumatic Brain Injury: A Pilot Study. *Frontiers in Neurology*, *12*, 598777. https://doi.org/10.3389/fneur.2021.598777

Roberts, S., Osborne, M., Ebden, M., Reece, S., Gibson, N., & Aigrain, S. (2013). Gaussian processes for time-series modelling. *Philosophical Transactions of the Royal Society A: Mathematical, Physical and Engineering Sciences*, *371*(1984), 20110550. https://doi.org/10.1098/rsta.2011.0550

Rubanova, Y., Chen, R. T., & Duvenaud, D. K. (2019). Latent ordinary differential equations for irregularly-sampled time series. *Advances in Neural Information Processing Systems*, *32*.

Schwedt, T. J. (2021). Post-traumatic headache due to mild traumatic brain injury: Current knowledge and future directions. *Cephalalgia*, *41*(4), 464–471. https://doi.org/10.1177/0333102420970188

Shan, S., Li, Y., & Oliva, J. B. (2023). Nrtsi: Non-recurrent time series imputation. *ICASSP 2023-2023 IEEE International Conference on Acoustics, Speech and Signal Processing (ICASSP)*, 1–5.

Tan, Z.-H., Sarkar, A. Kr., & Dehak, N. (2020). rVAD: An unsupervised segment-based robust voice activity detection method. *Computer Speech & Language*, *59*, 1–21. https://doi.org/10.1016/j.csl.2019.06.005

Tipirneni, S., & Reddy, C. K. (2022). Self-supervised transformer for sparse and irregularly sampled multivariate clinical time-series. *ACM Transactions on Knowledge Discovery from Data (TKDD)*, *16*(6), 1–17.

Weerakody, P. B., Wong, K. W., Wang, G., & Ela, W. (2021). A review of irregular time series data handling with gated recurrent neural networks. *Neurocomputing*, *441*, 161–178. https://doi.org/10.1016/j.neucom.2021.02.046

Williams, C., Klanke, S., Vijayakumar, S., & Chai, K. (2008). Multi-task gaussian process learning of robot inverse dynamics. *Advances in Neural Information Processing Systems*, *21*.

Williams, C., & Rasmussen, C. (1995). Gaussian processes for regression. *Advances in Neural Information Processing Systems*, *8*.

Wilson, A., & Adams, R. (2013). Gaussian process kernels for pattern discovery and extrapolation. *International Conference on Machine Learning*, 1067–1075.

Wu, C. J. (1983). On the convergence properties of the EM algorithm. *The Annals of Statistics*, 95–103.

Yao, F., Müller, H.-G., & Wang, J.-L. (2005). Functional Data Analysis for Sparse Longitudinal Data. *Journal of the American Statistical Association*, *100*(470), 577–590. https://doi.org/10.1198/016214504000001745

Zhang, P., Ren, P., Liu, Y., & Sun, H. (2022). Autoregressive matrix factorization for imputation and forecasting of spatiotemporal structural monitoring time series. *Mechanical Systems and Signal Processing*, *169*, 108718. https://doi.org/10.1016/j.ymssp.2021.108718

Zhu, C., Byrd, R. H., Lu, P., & Nocedal, J. (1997). Algorithm 778: L-BFGS-B: Fortran subroutines for large-scale bound-constrained optimization. *ACM Transactions on Mathematical Software*, *23*(4), 550–560. https://doi.org/10.1145/279232.279236

# Appendices

## Appendix A: Proof of Proposition 1

By Bayes' rule and the assumed independence of the joint posterior factorizes into two terms:

$$p\left(\mu_0, \mu_1 \middle| Y, Z, \Theta^{(r-1)}\right) \propto p\left(\mu_0 \middle| Y_0, \Theta^{(r-1)}\right) \cdot p\left(\mu_1 \middle| Y_1, \Theta^{(r-1)}\right)$$

We show the derivation for $\mu_0$, and the same argument applies to $\mu_1$. The prior for $\mu_0$ is $\mu_0 \sim N(m_0, K_{\theta_0^{(r-1)}})$, and the likelihood for each observation $y_i$, with $z_i = 0$ is $y_i \sim N\left(\mu_0, K_{\theta^{(r-1)}} + \left(\sigma^{(r-1)}\right)^2 I\right)$.

Additionally, we impose a smoothing penalty $\frac{1}{2}\mu_0^T R \mu_0$, where $R(\rho)$ is constructed from a second-difference operator to penalize the discrete curvature of the class-mean curve. Let $D_2$ denote the unscaled second-difference matrix:

$$D_2 = \begin{bmatrix} 1 & -2 & 1 & 0 & \cdots & 0 \\ 0 & 1 & -2 & 1 & \cdots & 0 \\ \vdots & \ddots & \ddots & \ddots & \ddots & \vdots \\ 0 & \cdots & 0 & 1 & -1 & 1 \end{bmatrix}.$$

For a smoothing-scale parameter $\rho > 0$, we define

$$D(\rho) = D_2/\rho^2, \qquad R(\rho) = D(\rho)^T D(\rho) = \rho^{-4} D_2^T D_2.$$

The negative log-posterior becomes:

$$L_0 = \frac{1}{2}(\mu_0 - m_0)^T \left(K_{\theta_0^{(r-1)}}\right)^{-1} (\mu_0 - m_0) + \frac{1}{2}\mu_0^T R \mu_0 + \frac{1}{2}\sum_{i:z_i=0} (y_i - \mu_0)^T \left( K_{\theta^{(r-1)}} + \left(\sigma^{(r-1)}\right)^2 I\right)^{-1} (y_i - \mu_0) + C_0,$$

where the constant term is $C_0 \in \mathbb{R}$. Expanding terms and grouping quadratic forms:

$$L_0 = \frac{1}{2}\mu_0^T \left(\left(K_{\theta_0^{(r-1)}}\right)^{-1} + R + n_0 \left(K_{\theta^{(r-1)}} + \left(\sigma^{(r-1)}\right)^2 I\right)^{-1}\right)\mu_0 - \mu_0^T \left(\left(K_{\theta_0^{(r-1)}}\right)^{-1} m_0 + \sum_{i:z_i=0} \left(K_{\theta^{(r-1)}} + \left(\sigma^{(r-1)}\right)^2 I\right)^{-1} y_i\right) + C_{1,}$$

where the constant term is $C_1 \in \mathbb{R}$. Therefore, the posterior is Gaussian:

$$p\left(\mu_0 \middle| Y_0, \Theta^{(r-1)}\right) = N\left(\widetilde{m}_0, \widetilde{K}_0\right),$$

where

$$\widetilde{m}_0 = \widetilde{K}_0 \left( K^{-1}_{\theta_0^{(r-1)}} m_0 + \sum_{i:z_i=0} \left( K_{\theta^{(r-1)}} + \left(\sigma^{(r-1)}\right)^2 I \right)^{-1} y_i \right),$$

$$\widetilde{K}_0 = \left( K^{-1}_{\theta_0^{(r-1)}} + R + n_0 \left( K_{\theta^{(r-1)}} + \left(\sigma^{(r-1)}\right)^2 I \right)^{-1} \right)^{-1},$$

In the same way, we obtain the posterior mean and covariance for $\mu_1$.

**Appendix B: Proof of Proposition 2**

*Moment calculations*

The first part is to derive $U_i$. By definition,

$$\begin{aligned} U_i &= \mathbb{E}\left[x_i^T \beta\right] \\ &= \beta_0 + \mathbb{E}\left[\beta_1^T \int_T \phi(t) f_i(t) dt\right] \end{aligned}$$

By Fubini's theorem, we can interchange integration and expectation:

$$\begin{aligned} \mathrm{U_i} &= \beta_0 + \beta_1^T \int_T \phi(t) \mathbb{E}[f_i(t)] dt \\ &= \beta_0 + \beta_1^T \int_T \phi(t) \mathbb{E}\left[\mu_{z_i}(t) + K_\theta^{(t^*,t_i)} \left(K_\theta^{(t_i,t_i)} + \sigma^2 I\right)^{-1} \left(y_i - \mu_{z_i}(t_i)\right)\right] dt \\ &= \beta_0 + \beta_1^T \int_T \phi(t) [\widetilde{m}_{z_i}(t) + K_\theta^{(t^*,t_i)} \left(K_\theta^{(t_i,t_i)} + \sigma^2 I\right)^{-1} \left(y_i - \widetilde{m}_{z_i}(t_i)\right)] dt \end{aligned}$$

Similarly, we can derive $\mathrm{V_i}$:

$$\begin{aligned} V_i &= Var\left[x_i^T \beta\right] \\ &= Var\left[\beta_1^T \int_T \phi(t) f_i(t) dt\right] \\ &= \beta_1^T \left( Var\left[\int_T \phi(t) f_i(t) dt\right] \right) \beta_1 \\ &= \beta_1^T \left( \iint_T \phi(t) \phi(t')^T Cov[f_i(t), f_i(t')] dt dt' \right) \beta_1 \end{aligned}$$

We show how to calculate the covariance term:

$$
\begin{aligned}
&Cov[f(t), f(t')] \\
&\qquad = Cov\left[\mu_{z_i}(t) + K_\theta^{(t^*,t_i)}\left(K_\theta^{(t_i,t_i)} + \sigma^2 I\right)^{-1}\left(y_i - \mu_{z_i}(t_i)\right), \mu_{z_i}(t') \right. \\
&\qquad \left. + K_\theta^{(t^*,t_i)}\left(K_\theta^{(t_i,t_i)} + \sigma^2 I\right)^{-1}\left(y_i - \mu_{z_i}(t_i')\right)\right] \\
&\qquad = \widetilde{K}_{z_i} - B_i\widetilde{K}_{z_i} - \widetilde{K}_{z_i}B_i^T + B\widetilde{K}_{z_i}B_i^T,
\end{aligned}
$$

where $B_i = K_\theta^{(t^*,t_i)}\left(K_\theta^{(t_i,t_i)} + \sigma^2 I\right)^{-1}$ and $\widetilde{K}_{z_i} = Cov\left[\mu_{z_i}(t), \mu_{z_i}(t')\right]$.

*Taylor series approximation*

Let $X = x_i^T\beta$. Denote $\mu_X = \mathbb{E}[X]$ and $\sigma_X^2 = Var[X]$. Note that $\mu_X = U_i$ and $\sigma_X^2 = V_i$. Let $g(X) = e^X$.

By the second-order Taylor series approximation around $\mu_X$, we have:

$$
\begin{aligned}
\mathbb{E}[g(X)] &\approx \mathbb{E}\left[g(\mu_X) + g'(\mu_X)(X - \mu_X) + \frac{1}{2}g''(\mu_X)(X - \mu_X)^2\right] \\
&= g(\mu_X) + \frac{1}{2}g''(\mu_X)\mathbb{E}[(X - \mu_X)^2] \\
&= \exp(\mu_X) + \frac{1}{2}\exp(\mu_X)\,\sigma_X^2,
\end{aligned}
$$

which becomes:

$$
\mathbb{E}\left[\exp\left(x_i^T\beta\right)\right] = \exp(U_i) + \frac{1}{2}\exp(U_i)\,V_i
$$

For the first-order approximation for $Var[\mathrm{g}(X)]$, we linearize $g(X)$ around $\mu_X$:

$$
\begin{aligned}
Var[g(X)] &\approx Var[g(\mu_X) + g'(\mu_X)(X - \mu_X)] \\
&= (g'(\mu_X))^2 Var[X] \\
&= \exp(2\mu_X)\,\sigma_X^2,
\end{aligned}
$$

which becomes:

$$
Var\left[\exp\left(x_i^T\beta\right)\right] \approx \exp(2U_i)\,V_i
$$

Let $h(X) = \log(1 + e^X)$. By a similar second-order argument for $\mathbb{E}[h(X)]$:

$$
\mathbb{E}[h(X)] \approx \log(1 + \mathbb{E}[\exp(X)]) - \frac{Var[\exp(X)]}{2(1 + \mathbb{E}[\exp(X)])^2}
$$

Substituting $\mathbb{E}[\exp(X)]$ and $Var[\exp(X)]$ yields:

$$\mathbb{E}[\log(1+\exp(X))] \approx \log\left(1+\exp(U_i)+\frac{1}{2}\exp(U_i)\,V_i\right) - \frac{\exp(2U_i)V_i}{2\left(1+\exp U_i+\frac{1}{2}\exp(U_i)V_i\right)^2}$$

**Appendix C: Proof of Proposition 3**

Let $X = x_i^T\beta$. Denote $\mu_X = \mathbb{E}[X]$ and $\sigma_X^2 = Var[X]$. Note that $\mu_X = U_i$ and $\sigma_X^2 = V_i$. Let $g(X) = e^X$.

By the conditions in Proposition 2:

$$\mathbb{E}[g(X)] = g(\mu_X) + \frac{1}{2}g''(\mu_X)\mathbb{E}[(X-\mu_X)^2] + \mathbb{E}\left[R_{g,3}\right],$$

The remainder in the mean approximation becomes:

$$\begin{aligned}\mathbb{E}\left[R_{g,3}\right] &= \frac{1}{6}\mathbb{E}[g'''(\xi)(X-\mu_X)^3] \\ &\leq \frac{M}{6}\mathbb{E}[|(X-\mu_X)^3|],\end{aligned}$$

for some $\xi$ in the interval between $X$ and $\mu_X$ and a constant $M$ such that $|g'''(\xi)| \leq M < \infty$.

*Sub-Gaussian moment bounds*

By the moment-generating function of X with variance parameter $\gamma^2 \geq \sigma_X^2$ for all real $\lambda$,

$$\mathbb{E}\left[\exp\left(\lambda(X-\mu_X)\right)\right] \leq \exp\left(\frac{\gamma^2\lambda^2}{2}\right).$$

From Markov's inequality, for any $t > 0$,

$$\begin{aligned}P(X-\mu_X \geq t) &= P\left(\exp\left(\lambda(X-\mu_X)\right) \geq \exp(\lambda t)\right) \\ &\leq \frac{\mathbb{E}\left[\exp\left(\lambda(X-\mu_X)\right)\right]}{\exp(\lambda t)} \\ &\leq \frac{\exp\left(\frac{\gamma^2\lambda^2}{2}\right)}{\exp(\lambda t)}\end{aligned}$$

By choosing $\lambda = \frac{t}{\gamma^2}$,

$$P(|X-\mu_X| \geq t) \leq 2\exp\left(-\frac{t^2}{2\gamma^2}\right)$$

We use the integral form of the $k$-th moment:

$$
\begin{aligned}
\mathbb{E}\left[\left|(X-\mu_X)^k\right|\right] &= \int_0^\infty P\left(\left|(X-\mu_X)^k\right| \geq t\right) dt \\
&= \int_0^\infty P\left(|X-\mu_X| \geq t^{1/k}\right) dt \\
&\leq \int_0^\infty 2\exp\left(-\frac{t^{2/k}}{2\gamma^2}\right) dt \\
&= (2\gamma^2)^{k/2} k \int_0^\infty e^{-u} u^{k/2-1}\, du \\
&= (2\gamma^2)^{k/2} k \Gamma(k/2) \\
&= O\left(\gamma^k\right)
\end{aligned}
$$

Since $\gamma^2 \geq \sigma_X^2$ and for sub-Gaussian distributions $\gamma^2$ is proportional to the actual variance, we have:

$$
\begin{aligned}
\mathbb{E}\left[R_{g,3}\right] &\leq \frac{M}{6}\mathbb{E}[|(X-\mu_X)^3|] \\
&= O(\gamma^3) \\
&= O\left(\left(Var\left(x_i^T\beta\right)\right)^{3/2}\right)
\end{aligned}
$$

*Extension to variance approximation*

Similarly, for the Taylor expansion for $\mathbb{E}[g(X)^2]$:

$$
\mathbb{E}[g(X)^2] = g(\mu_X)^2 + \left([g'(\mu_X)]^2 + g(\mu_X)g''(\mu_X)\right)\gamma^2 + O(\gamma^3)
$$

Then, we calculate the first-order expansion for $Var[g(X)]$:

$$
\begin{aligned}
Var[g(X)] &= \mathbb{E}[g(X)^2] - (\mathbb{E}[g(X)])^2 \\
&= [g'(\mu_X)]^2\gamma^2 + O(\gamma^3)
\end{aligned}
$$

The remainder term is again $O(\gamma^3) = O\left(\left(Var\left(x_i^T\beta\right)\right)^{3/2}\right)$. Therefore, the remainder terms in the Taylor expansion of $\mathbb{E}[\log(1+\exp(X))]$ are also bounded by a constant multiple of $\left(Var\left(x_i^T\beta\right)\right)^{3/2}$.

**Appendix D: Proof of Proposition 4**

By Proposition 2, $Var(x_i^T\beta)$ is given by an integral of the form:

$$
Var\left(x_i^T\beta\right) = \beta_1^T\left(\iint_T \phi(t)\phi(t')^T Cov[f_i(t), f_i(t')]dtdt'\right)\beta_1
$$

Since $Cov[f_i(t), f_i(t')]$ is decomposed in terms of $B$ and $\widetilde{K}_{z_i}$, both of which are constructed from RBF kernels, the covariance function is uniformly bounded. Additionally, because the basis function is bounded $\|\phi(t)\| \leq 1$, there exists a finite constant $M_{cov}$ such that:

$$\begin{aligned}\left\|Var(x_i^T\beta)\right\| &\leq \beta_1^T\left(\iint_T \|\phi(t)\|\|\phi(t')\|^T\|Cov[f_i(t), f_i(t')]\|dtdt'\right)\beta_1 \\ &\leq M_{cov} \cdot |T|^2\|\beta_1\|_2^2,\end{aligned}$$

Hence, we have

$$Var\left(x_i^T\beta\right) = O(\|\beta_1\|_2^2).$$

**Appendix E: Proof of Proposition 5**

To show that $y_{new}$ has this Gaussian distribution, we first compute the conditional mean:

$$\begin{aligned}\mathbb{E}\left[y_{new}|Y_{z_{new}}\right] &= \int y_{new}p\left(y_{new}|Y_{z_{new}}\right)dy_{new} \\ &= \int y_{new}\int p\left(y_{new}|\mu_{z_{new}}\right)p\left(\mu_{z_{new}}|Y_{z_{new}}\right)d\mu_{z_{new}}\,dy_{new} \\ &= \int\left[\int y_{new}p\left(y_{new}|\mu_{z_{new}}\right)dy_{new}\right]p\left(\mu_{z_{new}}|Y_{z_{new}}\right)d\mu_{z_{new}} \\ &= \int \mu_{z_{new}}p\left(\mu_{z_{new}}|Y_{z_{new}}\right)d\mu_{z_{new}} \\ &= \widetilde{m}_{z_{new}}\end{aligned}$$

Next, we compute the second moment:

$$\begin{aligned}\mathbb{E}\left[y_{new}^2|Y_{z_{new}}\right] &= \int y_{new}^2p\left(y_{new}|Y_{z_{new}}\right)dy_{new} \\ &= \int y_{new}^2\int p\left(y_{new}|\mu_{z_{new}}\right)p\left(\mu_{z_{new}}|Y_{z_{new}}\right)d\mu_{z_{new}}\,dy_{new} \\ &= \int\left[\int y_{new}^2p\left(y_{new}|\mu_{z_{new}}\right)dy_{new}\right]p\left(\mu_{z_{new}}|Y_{z_{new}}\right)d\mu_{z_{new}} \\ &= \int\left[K_{\widehat{\Theta}} + \sigma^2I + \mu_{z_{new}}^2\right]p\left(\mu_{z_{new}}|Y_{z_{new}}\right)d\mu_{z_{new}} \\ &= K_{\widehat{\Theta}} + \sigma^2I + \int \mu_{z_{new}}^2p\left(\mu_{z_{new}}|Y_{z_{new}}\right)d\mu_{z_{new}} \\ &= K_{\widehat{\Theta}} + \sigma^2I + \widetilde{K}_{z_{new}} + \widetilde{m}_{z_{new}}^2\end{aligned}$$

The conditional variance becomes:

$$Var\left[y_{new}|Y_{z_{new}}\right] = \mathbb{E}\left[y_{new}^2|Y_{z_{new}}\right] - \left(\mathbb{E}\left[y_{new}|Y_{z_{new}}\right]\right)^2$$
$$= K_{\widehat{\Theta}} + \sigma^2 I + \widetilde{K}_{z_{new}}$$

Therefore,

$$p\left(y_{new}(t)\middle|z_{new}, Y, \widehat{\Theta}\right) = N\left(\widetilde{m}_{z_{new}}(t), \widetilde{K}_{z_{new}}\right),$$

where $\tilde{\Sigma}_{z_{new}} = \widetilde{K}_{z_{new}} + K_{\widehat{\Theta}} + \sigma^2 I$.